\documentclass[runningheads]{llncs}
\usepackage[T1]{fontenc}
\usepackage{graphicx}

\usepackage[linesnumbered,ruled,vlined]{algorithm2e}

\usepackage{amsmath}
\usepackage{amsfonts}
\usepackage{amssymb}
\usepackage{bm}
\usepackage{siunitx}
\usepackage{calc}
\usepackage{setspace}

\usepackage{graphicx}  
\usepackage{caption}  
\usepackage{subcaption}  
\usepackage{float}
\usepackage{xcolor}
\usepackage{multirow}
\usepackage{multicol}
\usepackage{tabularx}
\usepackage{makecell}

\usepackage{comment}
\usepackage{cite}

\usepackage{tikz}

\usepackage{xspace}

\definecolor{lightblue}{rgb}{0., 0.1, 1.0}
\definecolor{sdp1}{rgb}{1.0, 0.16, 0.}
\definecolor{sdp2}{rgb}{1.0, 0.69, 0.}
\newcommand{\SDP}{\textit{Sob+Diff}\xspace}
\newcommand{\SDPcolor}{\textcolor{sdp1}{\SDP}\xspace}

\definecolor{dp1}{rgb}{0.0, 0.53, 0.01}
\definecolor{dp2}{rgb}{0.73, 0.93, 0.01}
\newcommand{\DP}{\textit{Diff}\xspace}
\newcommand{\DPcolor}{\textcolor{dp1}{\DP}\xspace}

\definecolor{pddp1}{rgb}{0.60, 0.62, 0.23}
\definecolor{pddp2}{rgb}{0.28, 1.0, 0.56}
\newcommand{\PDDP}{\textit{MLP}\xspace}
\newcommand{\PDDPcolor}{\textcolor{pddp1}{\PDDP}\xspace}

\definecolor{pddps1}{rgb}{0.0, 0.25, 1.0}
\definecolor{pddps2}{rgb}{0.11, 0.62, 1.0}
\newcommand{\PDDPS}{\textit{Sob+MLP}\xspace}
\newcommand{\PDDPScolor}{\textcolor{pddps1}{\PDDPS}\xspace}

\definecolor{oc1}{rgb}{0.62, 0.16, 0.71}
\definecolor{oc2}{rgb}{1.0, 0.43, 0.67}
\newcommand{\OC}{\textit{TO alone}\xspace}
\newcommand{\OCcolor}{\textcolor{oc1}{\OC}\xspace}

\newcommand{\SDPloss}{$\bm{\mathcal{L}}^{\textbf{\textit{Sob+Diff}}}$}
\newcommand{\npl}{n_{\textit{pl}}}
\newcommand{\ntraj}{n_{\textit{traj}}}
\newcommand{\nalgo}{n_{\textit{algo}}}

\newcommand{\datadistrib}{p_{\mathcal{D}}}
\newcommand{\ohist}{o_{\textit{hist}}}

\newcommand{\alphasob}{\alpha_{\textit{Sob}}}

\newcommand{\UR}[2]{%
    \textit{UR5($r_{\textit{init}}\!\!=\!\!#1$,$r_{\textit{tgt}}\!\!=\!\!#2$)}%
}
\newcommand{\subfigureplot}[4]{
    \begin{subfigure}[][][]{0.47\columnwidth}
        \begin{center}%
            \includegraphics[width=0.94\textwidth,trim={#4 #4 #4 #4},clip]{#1}%
            \vspace{-0.15cm}\caption{#2}%
            \label{#3}%
        \end{center}%
    \end{subfigure}%
}

\title{
Accelerating trajectory optimization with Sobolev-trained diffusion policies
}
\titlerunning{
Accelerating trajectory optimization with Sobolev-trained diffusion policies
}

\author{
    Théotime {Le Hellard}\inst{1,\star}\orcidID{0009-0008-4880-6341} \and 
    Franki {Nguimatsia Tiofack}\inst{1,\star}\orcidID{0009-0006-4227-2241} \and 
    Quentin {Le Lidec}\inst{1,2}\orcidID{0000-0001-7973-1030} \and 
    Justin Carpentier\inst{1}\orcidID{0000-0001-6585-2894}
}

\authorrunning{T. Le Hellard, F. Nguimatsia Tiofack, Q. Le Lidec and J. Carpentier}

\institute{
    Inria - Département d'Informatique de l'École normale supérieure, PSL Research University, France \email{\{firstname.surname\}@inria.fr} \and
    Courant Institute, New York University, USA \email{quentin.l@nyu.edu}
}

\begin{document}

\maketitle

\begin{abstract}
    Trajectory Optimization (TO) solvers exploit known system dynamics to compute locally optimal trajectories through iterative improvements. A downside is that each new problem instance is solved independently; therefore, convergence speed and quality of the solution found depend on the initial trajectory proposed. To improve efficiency, a natural approach is to warm-start TO with initial guesses produced by a learned policy trained on trajectories previously generated by the solver.
    Diffusion-based policies have recently emerged as expressive imitation learning models, making them promising candidates for this role. Yet, a counterintuitive challenge comes from the local optimality of TO demonstrations: when a policy is rolled out, small non-optimal deviations may push it into situations not represented in the training data, triggering compounding errors over long horizons.
    In this work, we focus on learning-based warm-starting for gradient-based TO solvers that also provide feedback gains. Exploiting this specificity, we derive a first-order loss for Sobolev learning of diffusion-based policies using both trajectories and feedback gains. Through comprehensive experiments, we demonstrate that the resulting policy avoids compounding errors, and so can learn from very few trajectories to provide initial guesses reducing solving time by $2\times$ to $20 \times$. Incorporating first-order information enables predictions with fewer diffusion steps, reducing inference latency.

    \keywords{Machine Learning in Robotics \and Control Theory and Optimization \and Diffusion Models}
\end{abstract}

\section{Introduction}
\renewcommand*{\thefootnote}{\fnsymbol{footnote}}
\footnotetext[1]{Equal contributions}
\renewcommand*{\thefootnote}{\arabic{footnote}}
\label{sec:introduction}
Given an instance of a control problem and an initially proposed sequence of states and controls, trajectory optimization (TO) methods iteratively refine the trajectory to meet problem constraints and minimize an associated cost function~\cite{jacobson1970differential,li2004iterative,posa2014direct}.
Being local methods, their solving time and the quality of the solution found directly depend on the initial trajectory. To complement TO with learning approaches, a natural application is to train a global policy from a batch of solved instances~\cite{levine2013GPS,levine2016GPSvisuomotor,mordatch2014combining,grandesso2023cacto}.
The resulting policy can then either be deployed or used to warm-start the solver on other instances.
In this paper, we study interplay loops alternating between collecting trajectories and training, close to DAgger~\cite{ross2011reduction}.
The policy provides initial guesses, the TO solver refines them towards locally optimal solutions, which are then used to train the policy.

For the imitation learning part, we leverage diffusion models\cite{sohl2015deep,ho2020denoising,song2020score} for their generalizing capacities.
Initially developed for image generation, diffusion-based policies~\cite{janner2022planning,ajay2022conditional,wang2022diffusion,chi2023diffusion} are now the favored approach in imitation learning, driving significant investments in robotics.
While commonly used to imitate human demonstrations, in this work, we use diffusion models to imitate trajectories generated by a solver~\cite{li2024diffusolve,yang2025physics}.

We focus on gradient-based TO solvers~\cite{jacobson1970differential,li2004iterative,mordatch2012discovery,jallet2025proxddp}, which leverage the derivatives of underlying dynamics algorithms~\cite{lee2005newton,giftthaler2017automatic,carpentier2018analytical} and the development of differentiable simulators~\cite{de2018end,le2021differentiable,lidec2024end}.
These solvers base their iterative steps on the computation of feedback gains, first-order information that can then be exploited at no additional cost~\cite{dantec2022first}.
In particular, they can help train a policy to imitate trajectories generated by gradient-based TO~\cite{mordatch2014combining,le2023enforcing, alboni2024cacto}, using derivatives matching~\cite{mitchell1992explanation,simard2002transformation,czarnecki2017sobolev,srinivas2018knowledge,pfrommer2022tasil}.
We use the Sobolev learning~\cite{czarnecki2017sobolev} terminology, a first-order supervised method that accelerates convergence and mitigates overfitting, reducing the number of trajectories needed for the policy to generalize.

We propose adapting the Sobolev learning formulation to diffusion-based policies, in connection with a gradient-based TO solver.
Our framework alternates between collecting trajectories using a gradient-based TO solver and first-order policy learning.
Compared to non-diffusion-based first-order policies~\cite{mordatch2014combining,le2023enforcing}, our method scales to more complex tasks, and compared to non-Sobolev methods, it requires fewer trajectories.
The resulting policy can predict over longer horizons, with fewer diffusion steps, hence accelerating diffusion inference, which in our context corresponds to the latency of predicting an initial guess.
Remarkably, it exhibits great resilience to the compounding error issue that imitation-based policies often struggle with, where small errors not observed during training cause the policy to fall out of distribution and enter a downward spiral.
In particular, we highlight a counterintuitive challenge when learning from TO-generated trajectories: their local optimality entails no small missteps, which might exacerbate the risk of compounding errors.
In fact, when training a diffusion policy from human demonstrations, it is common practice to include non-optimal, or even failing, trajectories, a recommendation opposite to learning from locally optimal trajectories.
Close to our work, \cite{li2024diffusolve} had to generate hundreds of thousands of trajectories in order to train diffusion policies from locally optimal trajectories.
In contrast, our Sobolev method benefits from the feedback gains, and only needs at most few hundred generated trajectories to provide effective initial guesses. \\[-0.5em]

In this work, we make the following contributions: \\[-1.7em]
\begin{itemize}
    \item we introduce a first-order loss for diffusion-based policy learning, to efficiently train with trajectories generated by gradient-based TO
    \item we propose an interplay algorithm, alternating between collecting trajectories and training, to solve hard tasks that TO struggles with without proper initial guesses,
    \item we present comprehensive experiments evaluating the sensitivity of (i) the number of collected trajectories, (ii) the number of training epochs, and (iii) the prediction horizon, across 3 different robotics tasks with 8 variants.
\end{itemize}
The rest is structured as follows.
Sec.~\ref{sec:background} provides background on trajectory optimization and diffusion, including prior works and challenges.
Sec.~\ref{sec:SDP} introduces our Sobolev approach to diffusion policies.
Sec.~\ref{sec:experiments} presents our set of experiments.
Finally, Sec.~\ref{sec:discussion} discusses related work, limitations, and future ideas.

\section{Background}
\label{sec:background}
\subsection{Optimal control problems}
We consider discrete-time dynamical systems. At time step $t$, we denote by $x_t$ the state of the system and $u_t$ the control input.
The system dynamic is defined by the differentiable function $f$, such that $x_{t+1}=f(x_t,u_t)$. In the context of robotics, $x_t$ is composed of joint positions $q_t$ and velocities $v_t$.
For indices $t_1 \le t_2$, we denote by $x_{t_1:t_2}$ the matrix whose rows are $x_{t_1}, x_{t_1+1}, \dots, x_{t_2}$.
The state and control trajectories are $X = x_{0:T}$ and $U = u_{0:T\!-\! 1}$.

We consider the following constrained optimal control problems~(OCPs):
\vspace{-0.5em}
\begin{equation}
    \begin{aligned}
        \underset{X, U }{\min} \quad
         & J(X, U ; \xi)
        = \sum_{t=0}^{T-1} \ell_t(x_t, u_t ; \xi) + \ell_T(x_T ; \xi) \\
        \text{s.t.} \quad
         & x_{t+1} = f(x_t,u_t), \quad  x_0 = \hat{x}(\xi),           \\
         & h_t(x_t, u_t ; \xi) \leq 0, \quad h_T(x_T ; \xi) \leq 0.
    \end{aligned}
    \label{OCP}
\end{equation}
$\xi$ denotes task parameters specifying the initial position $\hat{x}(\xi)$ and parameterizing both the differentiable constraints $h_t, h_T$ and the twice differentiable cost $J$. The latter is defined as the sum of stage costs $\ell_t$ and a terminal cost $\ell_T$.

\subsection{Gradient-based trajectory optimization}
\label{subsec:second-order-TO}

To efficiently find local optimum to OCPs, gradient-based TO solvers, such as iterative linear quadratic regulator (iLQR)~\cite{li2004iterative} and differential dynamic programming (DDP)~\cite {jacobson1970differential,jallet2025proxddp}, leverage the derivatives of stage costs, constraints, and of the dynamics function. In the case of robotic systems, these solvers exploit the derivatives of rigid body algorithms~\cite{giftthaler2017automatic,carpentier2018analytical} and differentiable physics simulators~\cite {de2018end,le2021differentiable,lidec2024end}.
The present paper uses iLQR-like methods, yielding local \textit{feedback gains} that we exploit in the policy learning part through Sobolev training.
In practice, we use the ProxDPP variant~\cite{jallet2025proxddp}, which handles constrained OCPs of the form~\eqref{OCP}.

iLQR iteratively optimizes a trajectory by (i) deriving a second-order Taylor expansion of the objective and first-order Taylor expansion of the dynamics near the current trajectory, and (ii) taking a step in the appropriate direction.
Let $J_t(x_{\geq t},u_{\geq t}) = \sum_{i \geq t}^{T\!-\! 1} \ell_i(x_i,u_i) + \ell_T(x_T)$ be the objective function from step $t$ onwards, $V_t(x_t) = \min_{x_{\geq t +1},u_{\geq t}} J_t$ the optimal value starting from a given $x_t$.
The Bellman principle states that $V_t(x_t) = \min_{u_t} \ell_t(x_t,u_t) + V_{t+1}(f_t(x_t,u_t))$, starting with $V_T(x_T) = \ell_T(x_T)$ and then $t$ from $T\!-\! 1$ down to $0$.
So, the Taylor expansion of $V_t$ around a nominal trajectory can be expressed using the second-order derivatives of the costs $\ell_t$ and dynamic $f_t$, and the Taylor expansion of the next value function $V_{t+1}$.
This expansion results in the computation of \textit{feedback gains} which correspond to $\frac{\partial u_t}{\partial x_t}$.

Thereupon convergence, in addition to the produced trajectories $X$ and $U$, these solvers estimate $\frac{\partial u_t}{\partial x_t}$, for all $t \!\in\!  [0,T\!-\! 1]$.
Then, as $x_{t+1} = f(x_t,u_t)$, by composing these derivatives with the differentiable dynamics, we get estimates of $\frac{\partial x_{t+1}}{\partial x_t}$ using the chain rule:
\vspace{-1em}
\begin{equation}
    \label{eq:chain}
    \frac{\partial x_{t+1}}{\partial x_t} =
    \frac{\partial f}{\partial x_t}(x_t,u_t)
    +
    \frac{\partial f}{\partial u_t}(x_t,u_t)
    \cdot
    \frac{\partial u_t}{\partial x_t}
\end{equation}

\subsection{Policy learning with diffusion models}
\label{subsec:diffusion}
\subsubsection{Related works.}
Policy learning trains policy $\pi_\theta$, with parameters $\theta \in \Theta$, to minimize the constrained objective function \eqref{OCP} in expectation over task parameters $\xi \sim \mathcal{P}$. Common Gaussian stochastic policies learn a conditional distribution of control inputs given the current state: $u_t \sim  \pi(u_t | x_t; \xi)$.
Recently, Janner et al~\cite{janner2022planning} instead proposed training diffusion models to control robots.

\textbf{Diffusion models}~\cite{sohl2015deep,ho2020denoising,song2020score,lipman2022flow} are probabilistic approaches that seek to generate elements from a desired distribution $\datadistrib$, given a training dataset \mbox{$\mathcal{D} = \{ \tau_0 \}$}.
Diffusion proceeds by progressively disrupting data through noise injection, then a model is trained to reverse this process for sample generation.
For diffusion policies~\cite{janner2022planning,ajay2022conditional,chi2023diffusion} $\mathcal{D}$ is built using successful trajectories of a desired control task.
Trajectories are sequences of states $x_t$ and/or controls $u_t$, with definitions varying across papers.
For unified notations, we denote by $a_t$ the action variable generated by the diffusion model, and by $o_t$ the observation variable used to condition the diffusion process, \textit{e.g.}, the current state.
When $a_t$ is $x_t$, or even $q_t$, $u_t$ is deduced by a low-level controller (PD control, IK solver, etc).

\textbf{Diffusion policies} are trained to model the distribution of trajectory chunks. Each training sample is a sequence $\tau_0 = a_{t_1:t_1+T_h-1}$, where $t_1$ is the start time and $T_h$ is the prediction horizon. For example, in robot control, $\tau_0$ may be a sequence of joint positions over the next $T_h$ time steps.
The model learns to generate a $T_h$-step control sequence conditioned on task parameters $\xi$ and the most recent $T_o$ states (including the current state), i.e., $o_{t-T_o+1:t}$, where $T_o$ is the history length. At inference time, the policy predicts a $T_h$-step sequence, executes the first $T_a$ steps, and then replans, where $T_a$ is the action length.

While diffusion models may be used as black box generative models, the present paper modifies their training loss by adding first-order information, so for completeness we detail an introduction to the diffusion framework used, denoising diffusion probabilistic models (DDPM)~\cite{ho2020denoising}.

\subsubsection{DDPM.}
A diffusion process is defined using two Markov chains with Gaussian transitions.
The \textit{forward noising} chain starts from \mbox{$\tau_0 \sim\datadistrib$} and gradually add noise over $K$ steps: \mbox{$\datadistrib(\tau_k | \tau_{k - 1}) := \mathcal{N}(\tau_k ; \sqrt{1 - \beta_k} \tau_{k - 1}, \beta_k \mathbf{I})$} for \mbox{$k \!\in\! [1,K]$}, with \mbox{$\beta_1, ..., \beta_K$} the noising schedule.
A sample $\tau_k$ can be directly drawn from $\tau_0$, by sampling a noise $\varepsilon \sim \mathcal{N}(0, \mathbf{I})$ and
\vspace{-0.5em}
\begin{equation}
    \label{eq:tauk}
    \tau_k \!:=\! \sqrt{\bar\alpha_k} \tau_0 + \sqrt{1 - \bar\alpha_k} \varepsilon
\end{equation}
with \mbox{$\bar \alpha_k \!:=\! \prod_{s=1}^k 1 \!-\! \beta_s$}.
Thus, forward posteriors conditioned on $\tau_0$ are tractable:
\begin{equation}
    \datadistrib(\tau_{k- 1} | \tau_k , \tau_0)
    = \mathcal N (\tau_{k- 1} ; \tilde \mu_k (\tau_k , \tau_0),
    \tilde \beta_k \mathbf I)
\end{equation}
\begin{equation}
    \label{eq:mu}
    \tilde \mu_k(\tau_k,\tau_0) =
    \tfrac{\sqrt{\bar \alpha_{k- 1}} \beta_k}{1 - \bar \alpha_k} \tau_0
    +
    \tfrac{\sqrt{\bar \alpha_k} (1 - \bar \alpha_{k- 1})}{1 - \bar \alpha_k} \tau_k
    \;\;
    {\displaystyle \text{and}}
    \;\;
    \tilde \beta_k = \tfrac{1 - \bar \alpha_{k- 1}}{1 - \bar \alpha_k} \beta_k
\end{equation}
In the other direction, the \textit{reverse} process starts from pure noise $\tau_K \!\sim\! \mathcal{N} (\mathbf{0} , \mathbf{I})$ and is trained to recover distribution $\datadistrib$ through a chain of Gaussians parameterized by $\theta$:
$p_\theta(\tau_{k\!-\! 1} | \tau_k) = \mathcal{N} (
    \tau_{k\!-\! 1} ;
    \mu_\theta (\tau_k , k),
    \tilde \beta_k \mathbf{I}
    )$ using the same $\tilde \beta_k$ as the forward posterior.
This way, the KL divergence between $\datadistrib$ and $p_\theta$ boils down to a weighted sum over $k \!\in\! [1, K]$ of the difference between $\tilde \mu_k(\tau_k , \tau_0)$ and $\mu_\theta (\tau_k , k)$. \\[-0.5em]

A neural network $\tau_\theta$ is trained to predict $\tau_0$ from $\tau_k$ and $k$, then at inference, sample $\tau_K \sim \mathcal{N}(0, \mathbf{I})$ and for $k$ from $K$ down to $1$, use the close form of $\tilde \mu_k$~\eqref{eq:mu} to sample $\tau_{k-1}$, \textit{i.e.}
$\tau_{k-1} \sim \mathcal{N}(\tilde \mu_k ( \tau_k , \tau_\theta(\tau_k,k)) , \tilde \beta_k \mathbf{I})$.
For diffusion policies, $\tau_0 = a_{t_1 : t_1 + T_h\!-\! 1}$, and $\tau_\theta$ is additionally conditioned on $\xi$ and $o_{\textit{hist}} = o_{t:t+T_o-1}$.
To further condition the generative process, \cite{janner2022planning} uses inpainting, overwriting \hbox{$\tau_k[1: T_o] = a_{\textit{hist}}$} at each denoising step, and if applicable $\tau_k[T_h] = x_{target}$ (a target position).
Finally, DDPM authors~\cite{ho2020denoising} recommend discarding the weighting when averaging over $k \!\in\! [1, K]$; using \eqref{eq:tauk}:
\begin{equation}
    \label{eq:ddpm}
    \mathcal{L}^{\textit{Diff}} =
    \underset{(\xi,\tau_0,o_{\textit{hist}}),\varepsilon,k}{\mathbb{E}}
    \left[
        \left\| \tau_0 - \tau_\theta (\tau_k , k, \xi, \ohist)\right\|^2
        \right]
\end{equation}
As an alternative parameterization, one can also choose to train $\varepsilon_\theta$ to predict the noise $\varepsilon$.
In our experiments, predicting $\tau_0$ performed better than predicting $\varepsilon$, so we will stick with the $\tau_0$ parametrization. Since $\tau_0$ can be deduced from $\varepsilon$ as $\tau_0 = \frac{1}{\sqrt{\bar \alpha_k}} (\tau_k - \sqrt{1 - \bar \alpha_k} \varepsilon)$, everything could be derived using $\varepsilon_\theta$ instead of $\tau_\theta$.

\subsubsection{The compounding error challenge.}
A bottleneck in diffusion-based policies is the number of trajectories needed.
In imitation learning, not having enough trajectories to cover the task space comes with the compounding error risk.
In particular, if the dataset does not contain missteps, small deviations during rollouts may lead the policy to unseen scenarios; without knowing how to recover, the policy may worsen the situation and enter a downward spiral.
To avoid this effect, the authors of diffusion policy~\cite{chi2023diffusion} later highlighted the need to include non-perfect trajectories in the dataset.
While it may turn the data collection process non-intuitive, having to predict how the policy could fail, similar to the DAgger method~\cite{ross2011reduction}, human teleoperated trajectories eventually bring enough diversity and imperfections.
In contrast, when using solvers for data collection, the local optimality may counterintuitively hurt.
In a pioneer work on diffusion for control, decision diffuser~\cite{ajay2022conditional} uses 10000 trajectories for the Kuka block stacking task.
Close to our work, DiffuSolve \cite{li2024diffusolve} uses about 200k locally optimal trajectories for a quadrotor navigation task.
While using a solver to generate demonstrations is usually cheaper, in \cite{li2024diffusolve}, collecting 200k trajectories took 20 hours on 200 CPU cores.
In this work, we will show how integrating the feedback gains by Sobolev learning mitigates this issue, exploiting the particularity of gradient-based TO.

\section{Interplay between gradient-based TO solvers and diffusion-based policies}
\label{sec:SDP}
We propose using Sobolev training to enhance diffusion policies trained on trajectories generated by gradient-based TO.
For complex tasks, our algorithm alternates between trajectory collection and training.

To recap all notations, $a_t$ is the action variable, either $x_t$, $u_t$ or a sub part, \textit{e.g.} $q_t$.
If $a_t$ is not $u_t$, the latter is estimated by inverse methods.
In our case, the observation variable $o_t$ includes $x_t$, $u_{t-1}$ and task parameters $\xi$.
Diffusion is done on chunks of variables: $\tau_0 = a_{t_1 : t_1 + T_h - 1}$.
The horizon $T_h$ is the length of this chunk, $T_a$ is the action length, \textit{i.e.} the number of applied actions before replanning, and $T_o$ is the history length, with $x_{\textit{hist}} = x_{t:t+T_o-1}$, $o_{\textit{hist}} = o_{t:t+T_o-1}$, $a_{\textit{hist}} = a_{t:t+T_o-1}$.
The first $T_o$ elements of $\tau_0$ are the current and past actions (for inpainting), only then come the $T_a$ used ones, hence $T_a \leq T_h - T_o$.

As explained Sec.~\ref{subsec:second-order-TO}, the TO solver produces feedback gains $\frac{\partial u_t}{\partial x_t}$, and by chaining with dynamics derivatives, we get $\tfrac{\partial x_{t+1}}{\partial x_t}$  \eqref{eq:chain}.
As $a_t$ is either $u_t$, $x_t$, or a subpart of it, we have $\frac{\partial a_t}{\partial x_t}$, and by further chaining these derivatives:
\vspace{-0.4em}
\begin{equation}
    \label{eq:bigchain}
    \tfrac{\partial a_{t+h}}{\partial x_{t+o}}
    = \tfrac{\partial x_{t+o+1}}{\partial x_{t+o}}
    \cdot \tfrac{\partial x_{t+o+2}}{\partial x_{t+o+1}}
    \cdot \dots
    \cdot \tfrac{\partial x_{t+h}}{\partial x_{t+h-1}}
    \cdot \tfrac{\partial a_{t+h}}{\partial x_{t+h}}
\end{equation}
for $t \!\in\! [0,T\!-\! 1]$ and $0 \! \leq \! o \! \leq h \! < \! T_h$.
Stacking derivatives gives  $\tfrac{\partial \tau_0}{\partial x_{\textit{hist}}} = \tfrac{\partial a_{t_1:t_1+T_h-1}}{\partial x_{t_1:t_1+T_o-1}}$.

\subsection{Sobolev learning for first-order diffusion policies}
In general, training a regression problem $f_\theta(x) = y$ in Sobolev spaces~\cite{czarnecki2017sobolev}, also known as derivatives matching~\cite{mitchell1992explanation,srinivas2018knowledge,pfrommer2022tasil} or tangent propagation~\cite{simard2002transformation}, consists of adding a first-order term in the regression loss, under a Frobenius norm:
\vspace{-0.8em}
\begin{equation}
    \label{eq:sobolev}
    \mathcal{L}^{\textit{Sobolev}} =\mathbb{E}_{x} \left[ \,\left\|f_\theta(x) - y\right\|^2
        + \alphasob
        \left\|\tfrac{\partial f_\theta(x)}{\partial x} - \tfrac{\partial y}{\partial x}\right\|^2_F
        \right]
\end{equation}
where $\frac{\partial y}{\partial x}$ is supposed known, and $\alphasob$ weights the first-order term.
Gradient descend of $\mathcal{L}^{\textit{Sobolev}}$ on $\theta$ requires computing the expensive $\partial_{\theta, x} f_\theta$ Hessian.
As an alternative, stochastic Sobolev training~\cite{czarnecki2017sobolev}
samples $n_{\textit{proj}}$ random vectors $v_i$ on the unit sphere (with $n_{\textit{proj}} =1$ turning out to be sufficient in our case), on which the gradients are projected:
\begin{equation}
    \label{eq:stochastic}
    \hspace{-0.3cm}
    \mathcal{L}^{\textit{Sobolev}}_{\textit{Stochastic}} = \mathbb{E}_{x} \left[ \,\left\|f_\theta(x) - y\right\|^2
    +
    \frac{\alphasob}{n_{\textit{proj}}}\sum\limits_{i=1}^{n_{\textit{proj}}}
    \left\| v_i^\top \cdot \frac{\partial y}{\partial x}
    - \frac{\partial}{\partial x}(v_i^\top \cdot f_\theta(x,\xi)) \right\|^2
    \right]
\end{equation}
\noindent \cite{mordatch2014combining} and \cite{le2023enforcing} respectively use \eqref{eq:sobolev} and \eqref{eq:stochastic}, to train a direct policy $\pi_\theta(x_t) = u_t$ using the feedback gains $\frac{\partial u_t}{\partial x_t}$.

One way to adapt \eqref{eq:sobolev} to diffusion policies could be to consider the whole generative process.
Comparing the original $\tau_0$ and its derivatives, to $\bar \tau^\theta_0$, the final output of the diffusion policy after all $K$ denoising steps, and the chained derivatives of all $K$ steps:
\vspace{-0.4em}
\begin{equation}
    \label{eq:sdp-full}
    \mathcal{L}^{\textit{Sob+Diff}}_{\textit{full}} =
    \mathbb{E}_{ (\xi,\tau_0,\frac{\partial \tau_0}{\partial x_{\textit{hist}}}),
    \tau_K}
    \left[ \left\| \tau_0 - \bar \tau^\theta_0 \right\|^2
        + \alphasob \left\| \tfrac{\partial \tau_0}{\partial x_{\textit{hist}}}
        - \tfrac{\partial \bar \tau^\theta_0}{\partial x_{\textit{hist}}}
        \right\|^2
        \right]
\end{equation}
However, this opposes how diffusion models are typically trained, and it did not work in practice.
Rather than training over the whole diffusion process at once, DDPM trains each denoising step separately, sampling $\varepsilon$ and $k$, and training $\tau_\theta$ to predict $\tau_0$ from $\tau_k$.
We name $\tau^{\theta,k}_0$ this estimate of $\tau_0$ from step $k$, not to be confused with $\bar\tau^\theta_0$, the latter being the final output of the whole diffusion process.
To appropriately combine $\mathcal{L}^{\textit{Diff}}$~\eqref{eq:ddpm} and $\mathcal{L}^{\textit{Sobolev}}_{\textit{Stochastic}}$~\eqref{eq:stochastic}, we propose to guide the derivatives of each denoising step:
\vspace{-0.3em}
\begin{small}
    \begin{equation}
        \label{eq:sdp}
        \bm{\mathcal{L}}^{\textbf{\textit{Sob+Diff}}}=
        \hspace*{-0.7cm}
        \underset{(\xi,\tau_0,\frac{\partial \tau_0}{\partial x_{\textit{hist}}}),
            \varepsilon, k}{\mathbb{E}}
        \Bigg[  \left\| \tau_0 - \tau^{\theta,k}_0\right\|^2
        + \frac{\alphasob}{n_{\textit{proj}}}
        \sum_{i=1}^{n_{\textit{proj}}}
        \left\| v_i^\top \cdot \frac{\partial \tau_0}{\partial x_{\textit{hist}}}
        - \frac{\partial}{\partial x_{\textit{hist}}}\left(v_i^\top \cdot \tau^{\theta,k}_0\right)\right\|^2
        \Bigg]%
    \end{equation}%
\end{small}%
Eq.~\eqref{eq:sdp} is not equivalent to Eq.~\eqref{eq:sdp-full}, as it steers the derivatives of each step toward the desired derivatives of the full process.
Moreover, the Sobolev weight $\alphasob$ is fixed for all diffusion steps.
This might be counterintuitive, as early steps ($k$ close to $K$) are stochastic, while late steps ($k$ close to 1) focus the distribution to one mode.
But oppositely, for $k < K$, at inference, $\tau_0^{\theta,k} = \tau_\theta(\tau_k^\theta,k,\xi,o_{\textit{hist}})$ has a dependence to $x_{\textit{hist}}$ through $\tau_k^\theta$, as it was obtained from a previous prediction.
While one could define $k$-dependent weights, \textit{e.g.} increasing $\alphasob^k \! = \! \alphasob\frac{k}{K}$ or decreasing $\frac{K - k}{K}$, we observed counteracting effects and let $\alphasob^k = \alphasob$.

\subsection{Alternating loop}
The full training procedure, Algorithm~\ref{algo:training}, repeats over $n_{\textit{algo}}$ iterations: (i) collect a dataset $\mathcal{D}$ of trajectories using TO ; (ii) train a policy on $\mathcal{D}$ using \SDPloss~\eqref{eq:sdp}; (i') collect new trajectories, but using the policy to provide initial guesses to the solver.
For policy rollout, Algorithm~\ref{algo:rollout} iteratively generates $T_h$ actions using the DDPM inference scheme, out of which $T_a$ actions are rolled out.
For simple tasks, collecting one fixed dataset $\mathcal D$ using TO is enough ($n_{\textit{algo}}=1$), but for harder ones, further collecting trajectories using the policy to provide initial guesses ($n_{\textit{algo}}>1$) helps reduce the solving time and can lead to better local minima.
This is due to \textbf{three key choices in Alg.~\ref{algo:training}}: Line~\ref{line:argmin-cost} calls the solver twice, both with and without warm-starting with the policy.
Line~\ref{line:reject} rejects trajectories where the solver failed to converge after $n_{\textit{max}}$ solving iterations.
Finally, Line~\ref{line:reset}, we chose to reset $\mathcal D$ between each algorithm iteration.
While more complex heuristics could be beneficial, through our experiments we found these designed choices to be effective.
On hard tasks, using TO alone may lead to high \textit{rejection}-rates (Line~\ref{line:reject}), so the policy reduces the total time needed to collect $n_{\textit{traj}}$ trajectories.

\begin{algorithm}
    \fontsize{8.5}{9.5}\selectfont
    \caption{Training - Interplaying collection and learning}
    \label{algo:training}

    \DontPrintSemicolon
    \SetKwInOut{Initialize}{Initialize}
    \Initialize{Diffusion policy network $\tau_\theta$; Buffer $\mathcal{D} = \emptyset$}

    \For{$n_{\textit{algo}}$
    }{
    \tcp{\textcolor{lightblue}{Collect new trajectories}}
    \textbf{if} \textit{reset buffer} \textbf{then} $\mathcal{D} = \emptyset$ \label{line:reset} \\
    \For{$n_{\textit{traj}}$}{
        Sample $\xi \sim \mathcal{P} \label{line:sample}$\\
        $X^{0}, U^{0} = \textit{Interpolate}(x_0,x_{\textit{target}})$ \\
        $X^{\pi}, U^{\pi}
            = \textbf{PolicyRollout}(\tau_\theta, \xi, T_a, K_{\textit{rollout}})$ \label{line:rollout}\\
        $X, U, \frac{\partial u_t}{\partial x_t}, \frac{\partial x_{t+1}}{\partial x_t}
            = \textit{ArgminCost} \:(
            \textbf{TO}(X^{0}, U^{0}, \xi), \:
            \textbf{TO}(X^{\pi}, U^{\pi}, \xi))
        $ \label{line:argmin-cost}\\
        \textbf{if} both \textbf{TO} calls did not converge in $n_{\textit{max}}$ iterations \\
        \textbf{then} \textit{reject - restart Line~\ref{line:sample}} \label{line:reject} \\
        \textbf{else}
        $\mathcal{D} \leftarrow \mathcal{D} \cup \{(X, U, \frac{\partial u_t}{\partial x_t}, \frac{\partial x_{t+1}}{\partial x_t})\}$
    }
    \BlankLine
    \tcp{\textcolor{lightblue}{Policy learning}}
    \For{$n_{\textit{pl}}$ epochs}{
    \tcp{\textcolor{lightblue}{Sample by batches}}
    Sample $\xi$, $\tau_0$, $x_{\textit{hist}}$, $o_{\textit{hist}}$ and derivatives in $\mathcal{D}$\\
    $\frac{\partial \tau_0}{\partial x_{\textit{hist}}}
        = \textit{ChainRule}(\frac{\partial x_{t+1}}{\partial x_t},
        \frac{\partial u_t}{\partial x_t})$ \label{line:chainrule} \tcp*{using \eqref{eq:bigchain}}
    $k \sim \mathcal{U}(1, K_{\textit{train}})$, $\; \varepsilon \sim \mathcal{N}(0, I)$\\
    \BlankLine
    \tcp{\textcolor{lightblue}{Apply noise and inpainting}}
    $\tau_k = \sqrt{\bar \alpha_k} \tau_0
        + \sqrt{1 - \bar \alpha_k} \varepsilon$ \\
    $\tau_k[1:T_o] = a_{\textit{hist}}$

    \BlankLine
    \tcp{\textcolor{lightblue}{Compute loss ${\mathcal{L}}^{{\textit{Sob+Diff}}}$~\eqref{eq:sdp}}}
    $\tau^{\theta,k}_0 = \tau_\theta(\tau_k,k,\xi,o_{\textit{hist}})$ \\
    Sample $n_{\textit{proj}}$ vectors $v_i$ in the unit sphere\\
    Take gradient descent step on $\nabla_\theta \mathcal{L}^{\textit{Sob+Diff}}(\theta)$
    }
    }
\end{algorithm}

\begin{algorithm}
    \fontsize{8.5}{9.5}\selectfont
    \caption{Policy Rollout}
    \label{algo:rollout}
    \SetKwInOut{Initialize}{Initialize}
    \DontPrintSemicolon
    \KwIn{$\tau_\theta$, $\xi$, $T_a$, $K_{\textit{rollout}}$}
    \Initialize{
        $t=0$, $o_{\textit{hist}}$, $a_{\textit{hist}}$
    }

    \While{$t < T$}{
    \tcp{\textcolor{lightblue}{Full reverse denoising process}}

    $\tau_K \sim \mathcal{N}(0, I)$\\

    \For{$k = K_{\textit{rollout}}$ \KwTo $1$}{
        $\tau_k[1:T_o] = a_{\textit{hist}}$ \tcp*{inpainting}
        $\tau^{\theta,k}_0 = \tau_\theta(\tau_k,k,\xi,o_{\textit{hist}})$ \\
        $\tau_{k-1} \sim \mathcal N(\tilde \mu_k(\tau_k,\tau^{\theta,k}_0), \tilde \beta_k \mathbf I)$
    }

    \BlankLine
    \tcp{\textcolor{lightblue}{Play predicted actions}}
    \For{$s = 1$ \KwTo $T_a$}{
    \textbf{if} $t+s > T$ \textbf{then \textit{stop}} \\
    \textbf{if} $a$ is $u$ \textbf{then} $u_{t+s-1} = \tau_0[s]$ \\
    \textbf{else} $u_{t+s-1} = \textit{Inverse}(x_{t+s-1}, \tau_0[s])$ \label{line:inverse} \tcp*{e.g. PD control, RNEA etc}
    $x_{t+s} = f(x_{t+s-1},u_{t+s-1})$ \tcp*{dynamics}
    }
    Update $o_{\textit{hist}}$ and $a_{\textit{hist}}$ \\
    $t = t + T_a$
    }
    \KwOut{$X$, $U$}
\end{algorithm}

\section{Experiments}
\label{sec:experiments}
In this section, we refer to our method as \SDP, and evaluate it against alternative policy learning architectures.
Projected DDP (\textit{PDDP}) and its Sobolev variant (\textit{PDDP+S}) from \cite{le2023enforcing} use a similar interplay loop and correspond to using a direct MLP (multilayer perceptron) instead of a diffusion model.
For simplicity, we will refer to PDDP and PDDP+S as \PDDP and \PDDPS, respectively.
\textit{DiffuSolve}~\cite{li2024diffusolve} trains diffusion policies over trajectories generated by a solver.
While setup and implementation differ (in particular, they do not use a gradient-based solver), DiffuSolve is close to ablating the Sobolev component, which we refer to by \DP in this section. DiffuSolve does not have an interplay loop to further collect trajectories, but we unify the name of the method with the learning components used.
The experiments are separated into two parts. $\nalgo$ is first fixed to 1 (no further collecting trajectories) to study how many trajectories are needed for the different policies to learn.
Then, for harder tasks, we study the full algorithm ($\nalgo > 1$), alternating between trajectory collection and policy learning, as described in Alg~\ref{algo:training}.

\subsection{Experimental setup and implementation details}
Our code will be made public upon paper acceptance.
We use Pinocchio~\cite{carpentier2019pinocchio} and \textsc{Aligator}~\cite{aligatorweb} to define and solve the constrained OCPs~\eqref{OCP}.
For the learning part, we use PyTorch with the same conditional U-Net as  \cite{janner2022planning} and \cite{chi2023diffusion}, the DDPM scheduler is square cosine from improved DDPM~\cite{nichol2021improved} and HuggingFace implementation.
All experiments were run on a single laptop, with an Intel Core i9-13950HX and an Nvidia RTX2000 ADA.
The number of iterations the TO solver can do is restricted to $n_{\textit{max}}=1000$.
While we find it beneficial to reset $\mathcal{D}$ after each iteration (Alg~\ref{algo:training}, Line~\ref{line:reset}), still, as the policy improves, we may choose to increase $\ntraj$, since trajectories are faster to collect (as the rejection-rate Line~\ref{line:reject} decreases).
We let $n_{\textit{proj}} = 1$, sampling only one vector for the stochastic Sobolev training, so training \SDP on one epoch takes about twice as much time as \DP.
We fix the hidden dimensions of the conditional U-Net to $[24, 24, 32, 32]$, resulting in approximately 300k parameters.
We fixed the number of diffusion steps $K$ to 5.
By default, $T_h = 32$, $T_o=1$ and $T_a = T_h - T_o = 31$.

To compare the training time fairly, we adapt the size $E$ of an "\textit{epoch}" to the method.
For direct policies, predicting $u_t$ given $x_t$ (\textit{e.g.}, \PDDP), $E = |\mathcal{D}| \times T$, the total number of state-action pairs in the dataset.
But for diffusion-based policies, predictions are made over horizon length $T_h$, covering $T_h$ steps, so we chose to define \mbox{$E = |\mathcal{D}| \times (T - T_h) / T_h$}, which is the number of chunks.

All results are averaged over 5 random seeds, displaying confidence intervals.
On each seed, the mean trajectory cost $J(X,U;\xi)$ is estimated using 50 random instances $\xi \!\sim\! \mathcal P$.
When the mean cost is higher than $10^5$, we put an $\infty$ sign at the top of the plot.
We assigned colors to methods for visual consistency across figures: \SDPcolor (red), \DPcolor (green), \PDDPcolor (gold), and \PDDPScolor (blue). As a reference, we evaluate the performance of TO with interpolation-based initial guesses, reported as \OCcolor (purple).

\subsection{Tasks descriptions}

Methods are evaluated on goal-reaching tasks, with a UR5, a quadrotor, and inverted pendulums.
The implementation is modular and works on any problem defined using \textsc{Aligator}~\cite{aligatorweb}.
We use $\alphasob =1.0$, except for the quadrotor task where $\alphasob =10^{-3}$ performs better. \\[-0.5em]

\noindent \textbf{Inverted pendulums.}
The goal consists of swinging a single or double pendulum from a randomly sampled downward position to the upright unstable equilibrium point.
To avoid ill-conditioned solutions, $|u_t|$ is bounded by 25 and a regularization on $u$ makes the task harder for the TO solver, $\ell_t(x_t,u_t) = 10 \left\| \textit{end-effector}(x_t) - \textit{goal}\right\|^2 + 0.1 \left\| u_t \right\|^2$.
The chaotic properties of this task impose torque control, $a_t=u_t$.\\[-0.5em]

\noindent \textbf{UR5.}
To gradually increase the complexity of the goal-reaching task with a UR5 robotic arm, we tried 4 variants.
We define two boolean variables: \textit{fully random init} $r_{\textit{init}}$, whether the initial position is sampled in a small reasonable area ($r_{\textit{init}}\!\!=\!\!0$) or anywhere in the robot space ($r_{\textit{init}}\!\!=\!\!1$); and \textit{random target} $r_{\textit{tgt}}$, whether the end-effector target position is fixed ($r_{\textit{tgt}}\!\!=\!\!0$) or random ($r_{\textit{tgt}}\!\!=\!\!1$).
As a reference, in \cite{le2023enforcing}, PDDP+S (\textit{i.e.} \PDDPS) was tested on \UR{0}{0}.
$q_t$ is composed of the 6 joint angles, and following~\cite{le2023enforcing}, $q_t$ is preprocessed to $(\cos \, q_t, \sin \, q_t)$, derivatives are adjusted accordingly (Alg~\ref{algo:training}, Line~\ref{line:chainrule}).
For this task, we set $a_t = x_t$, and $u_t$ is computed using RNEA~\cite{carpentier2019pinocchio} for Alg~\ref{algo:training} Line~\ref{line:inverse}, so only $v_t$ is used, but predicting $q_t$ too stabilizes the training.
The only constraints are torque limits.\\[-0.5em]

\noindent \textbf{Quadrotor.}
We first evaluate all methods on controlling a quadrotor to go from a random initial position to a random goal, at the same height, with torque limits, ceiling, and floor constraints.
Then, up to 9 columns are added to the environment as obstacles for the quadrotor to avoid.
$q_t$ is the SE3 placement of the quadrotor, preprocessed by projecting it to the tangent space \mbox{$dq_t = q_t \ominus_{\textit{\tiny SE3}} q_{\textit{ref}}$}, and since state control works better on this task too, $a_t = (dq_t,v_t)$.

\subsection{Learning capacities and sample efficiency}
\label{subsec:expe-learning}

\begin{figure}
    \begin{center}
        \includegraphics[width=\columnwidth]{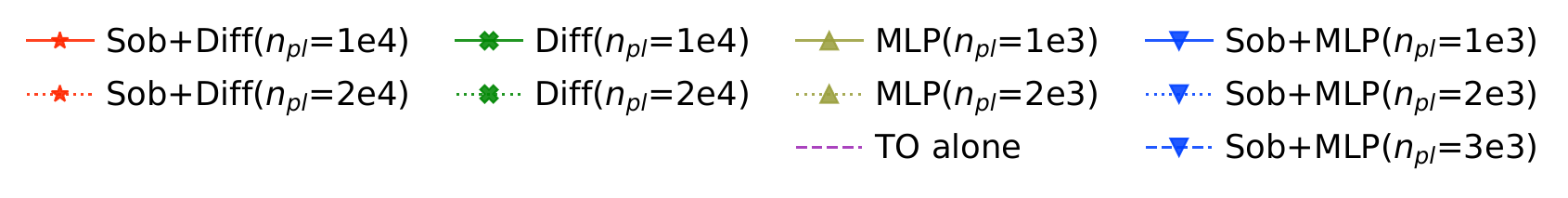}

        \subfigureplot{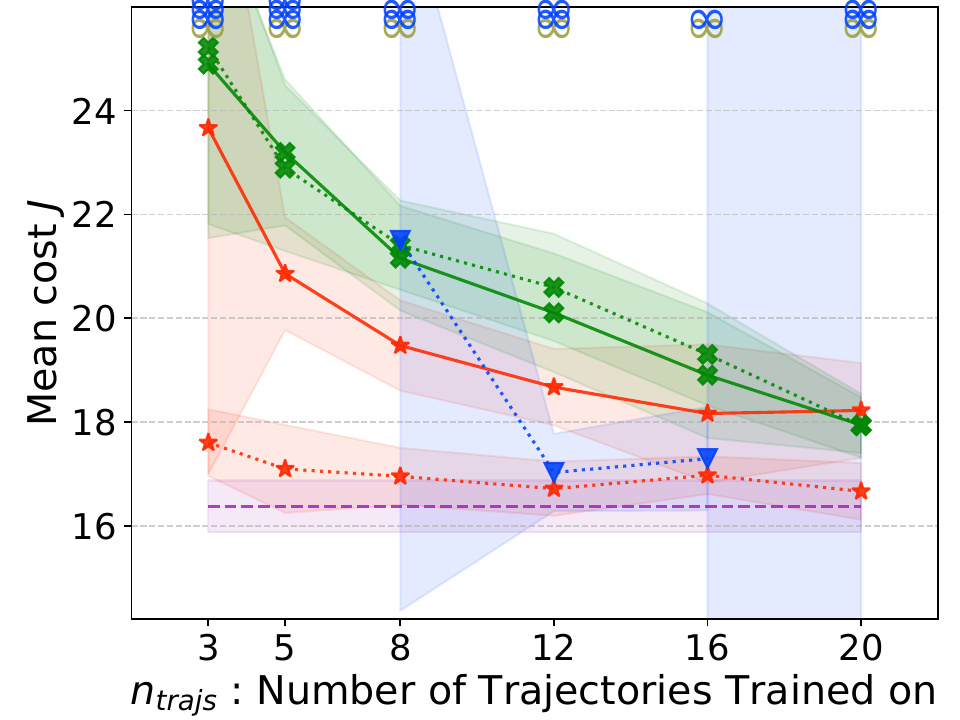}{\UR{0}{0}}{im:pl:UR00}{0}
        \hfill
        \subfigureplot{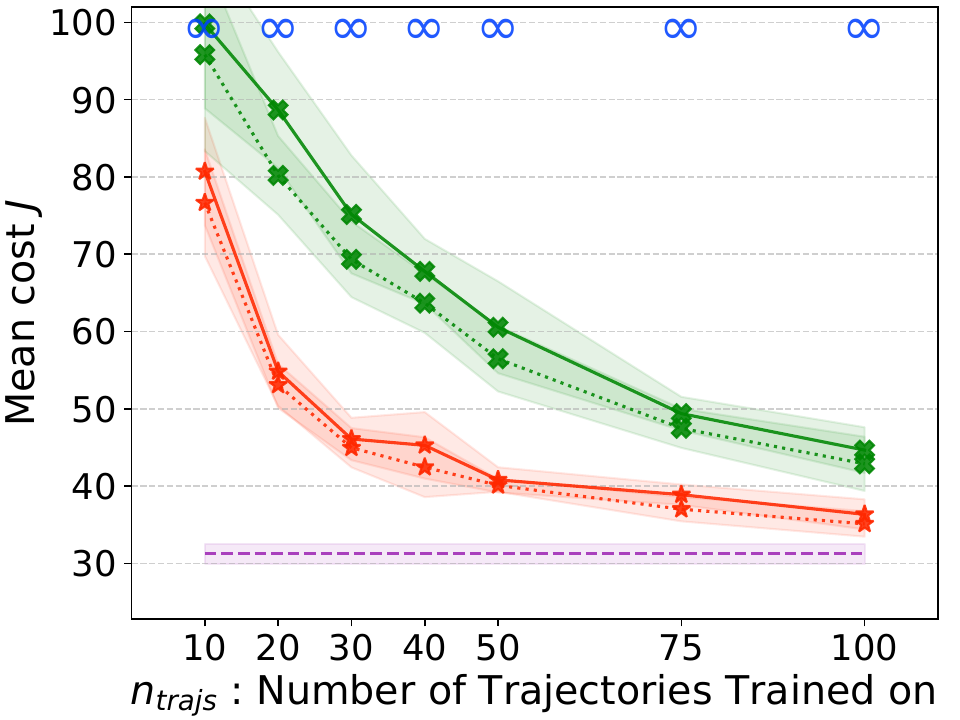}{\UR{0}{1}}{im:pl:UR01}{0}

        \vspace{0.35cm}
        \subfigureplot{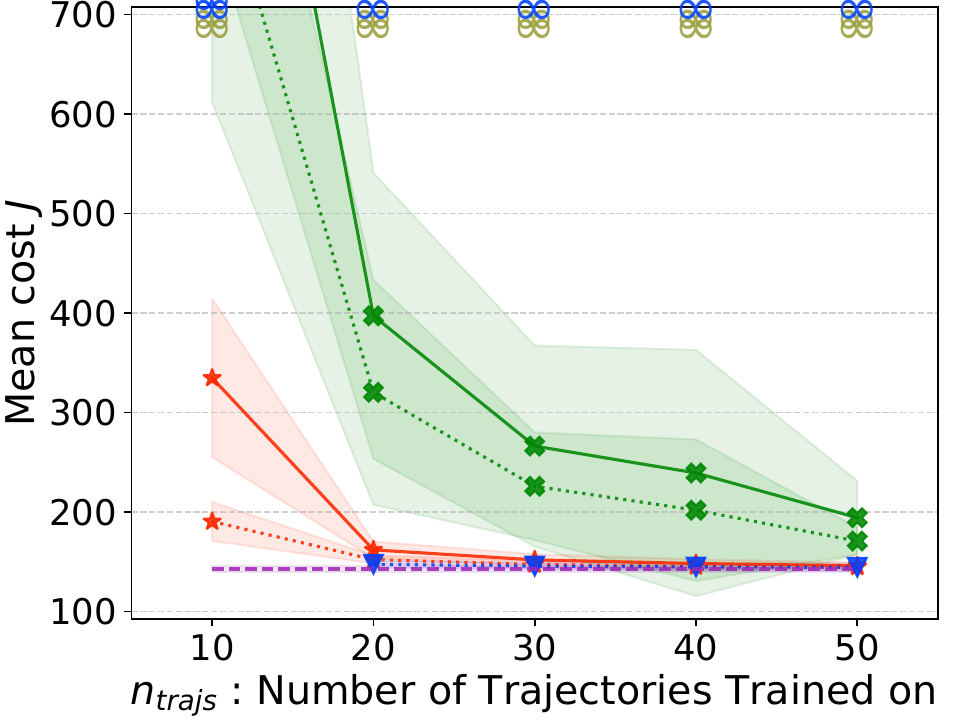}{Quadrotor without obstacles}{im:pl:quad}{0}
        \hfill
        \subfigureplot{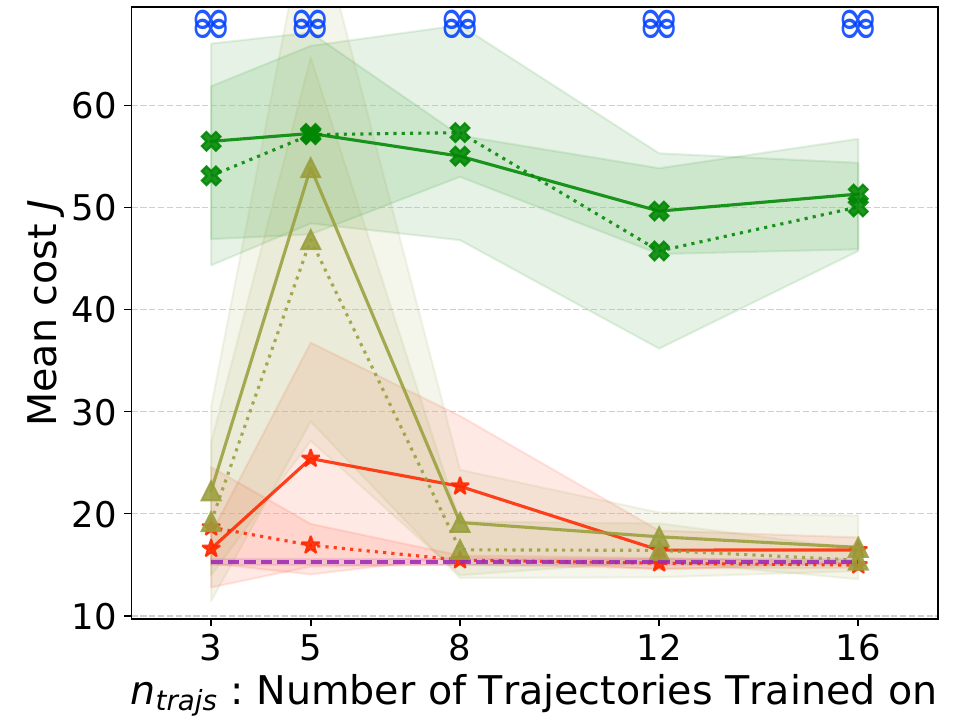}{Inverted single pendulum}{im:pl:pend}{0}

        \caption{\small \textbf{Policy learning capacities.} Plot mean cost $J(X,U;\xi)$ from \eqref{OCP} on test instances $\xi$ \textit{w.r.t} $\ntraj$, the number of trajectories in the dataset.
            Curves with the same color but different line styles differ in the number of training epochs $\npl$. \SDP reaches the performance of the TO solver on all tasks, even when trained with very few trajectories, while \DP needs between 5 to 10 times more trajectories, and both \PDDP and \PDDPS are very unstable. Our \SDP successfully operates the inverted pendulum with torque control (actually, on this task $\npl=10^3$ and $\npl=2\cdot10^3$).
        }
        \label{fig:pl-perfs}
    \end{center}
    \vspace{-1em}
\end{figure}

This subsection focuses on the learning capacities, $\nalgo$ is fixed to 1 (\textit{i.e.} no further trajectory collection), to see how many trajectories are needed for the policy to produce initial guesses with average costs close to typical ones the solver reaches without warm-starting.
On Fig.~\ref{fig:pl-perfs}, the sample efficiency is evaluated by plotting the average cost with respect to the dataset size $\ntraj = |\mathcal{D}|$. Results are collected independently, with a fixed dataset for each run, for example, one run had 5 trajectories, another one had 5 more etc.
To investigate the dependence on the number of training epochs $\npl$, each method is evaluated at different times during training.
Both \SDP and \DP are evaluated at $\npl=10^4$ and $\npl = 2 \cdot 10^4$ (except for the single pendulum, where $10^3$ is enough), while \PDDP and \PDDPS are evaluated at $\npl=10^3$ and $\npl=2 \cdot 10^3$.
On all tasks, \SDP reaches the average performance of the TO solver, achieving both state control and torque control, even when trained with very few trajectories.
Compared to \DP, \SDP usually needs between 5 to 10 times fewer trajectories.
On the inverted pendulum task, \SDP succeeds with $\ntraj=3$, while \DP systematically fails to operate torque control.
Overall, \PDDP and \PDDPS are unstable; they often diverge on some trajectories, which shifts their mean cost to extreme values.
On the \UR{0}{0} task, Fig.~\ref{im:pl:UR00}, \PDDPS is additionally evaluated with $\npl=3 \cdot 10^3$, it reveals that \PDDPS only succeeds with $8 \leq \ntraj \leq 16$ and $\npl = 2 \cdot 10^3$, otherwise following into overfitting.

\subsection{Full algorithm evaluation}
\label{subsec:exp-full}
\begin{figure}
    \begin{center}
        \includegraphics[width=0.8\columnwidth]{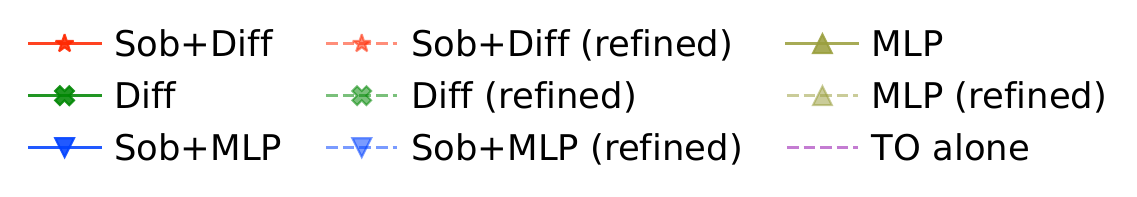}

        \subfigureplot{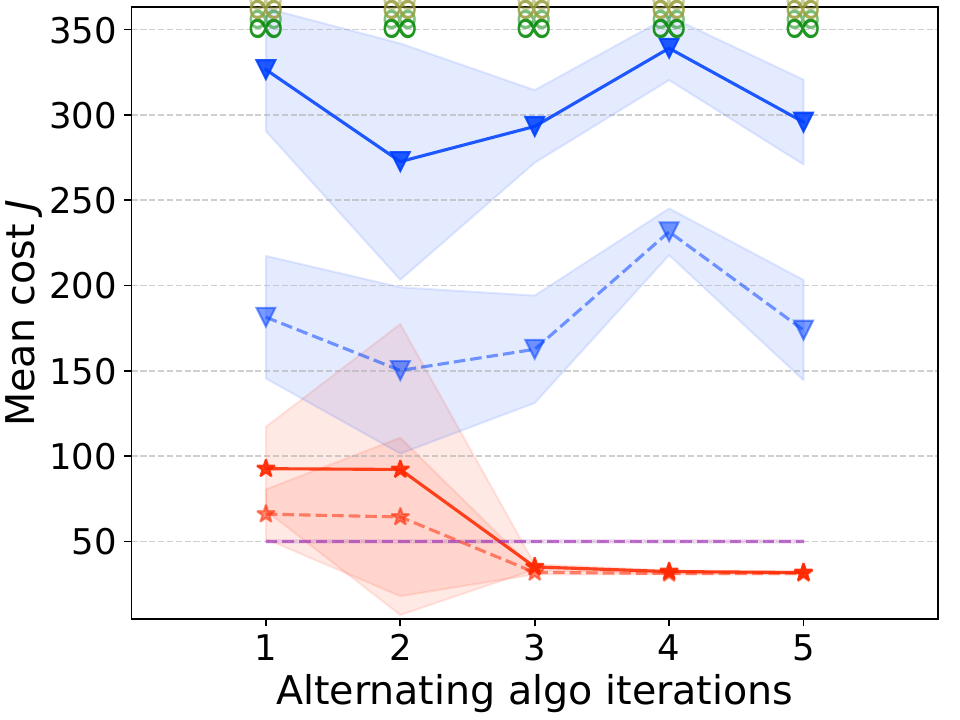}{Mean cost \textit{w.r.t.} algo iters, evaluating both the policy and \textit{refined}, \textit{i.e.} TO starting from the policy solution.}{im:pend-double}{0}
        \hfill
        \subfigureplot{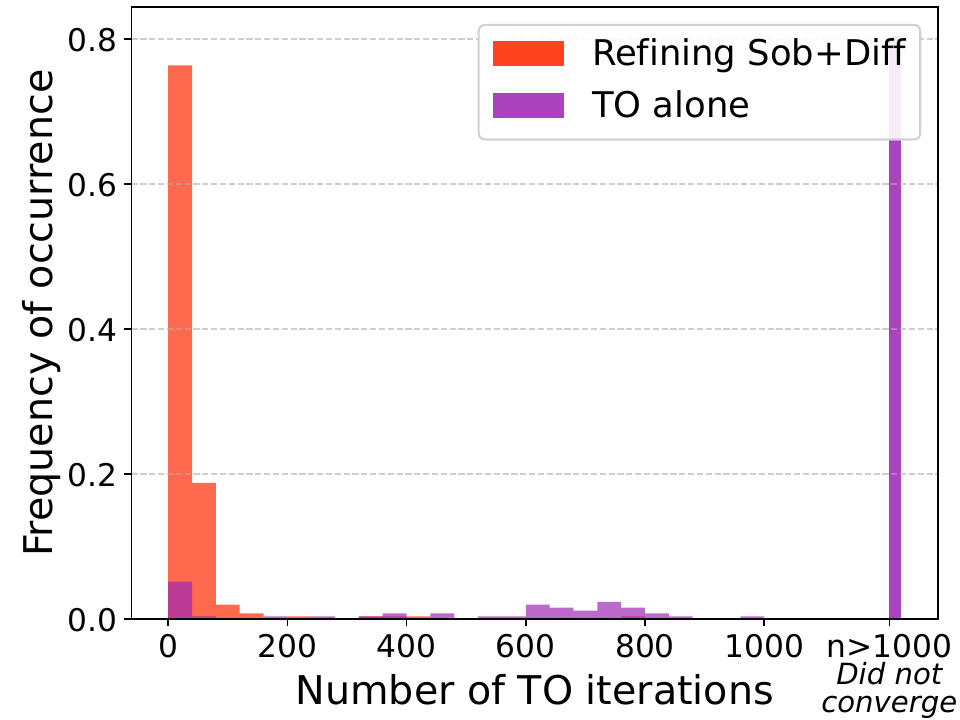}{Histogram of the number of TO iterations taken to converge, with \SDPcolor to warm-start and \textcolor{oc1}{without}.}{im:hist:pend-double}{0}

        \vspace{0.5em}
        \begin{center}
            \begin{minipage}{0.95\linewidth}
                \begin{subfigure}[][][]{\columnwidth}
                    \centering
                    \begin{tabular}{|c|c c c c c c c|c|}
                        \hline
                        $T_a$             & 1    & 2    & 4    & 6    & \textbf{9}   & 12   & 15   & TO Alone \\
                        \hline
                        Policy mean cost  & 31.7 & 31.7 & 31.8 & 32.5 & 33.5         & 85.1 & 137  & -        \\
                        Refined mean cost & 31.3 & 31.3 & 31.2 & 31.7 & 32.3         & 48.1 & 59.7 & 50.1     \\
                        Traj Opt iters    & 27   & 35   & 35   & 43   & 78           & 405  & 690  & 896      \\
                        Rollout time (s)  & 9.2  & 4.7  & 2.4  & 1.6  & 1.1          & 0.80 & 0.66 & -        \\
                        Solving time (s)  & 0.22 & 0.30 & 0.29 & 0.36 & 0.56         & 2.5  & 4.4  & 4.1      \\
                        Total time   (s)  & 9.4  & 5.0  & 2.6  & 1.9  & \textbf{1.6} & 3.3  & 5.1  & 4.1      \\
                        \hline
                    \end{tabular}
                    \caption{
                        \textbf{Performances at inference when varying $T_a$.}
                        The first two lines report the average trajectory cost $J$. The policy is the \SDP one obtained at the end of training from plot \textbf{(a)}.
                        The third line is the mean number of TO iterations taken to converge when refining (early stopped at 1000).
                        As a reference, the last column shows the cost and solving time when TO is not warm-started.
                        For our setup, $T_a=9$ appears to be a good trade-off between policy rollout time and TO solving time.
                    }
                    \label{tab:vary_a}
                \end{subfigure}
            \end{minipage}
        \end{center}

        \caption{\small \textbf{Interplay algorithm on the inverted double pendulum.}
            For this task, we use $T_h \!=\! 16$ and at training time $T_a \!=\! 4$ (Alg~\ref{algo:training}, Line~\ref{line:rollout}), $\ntraj\!=\!30$ and $\npl \!=\! 3 \cdot 10^3$ for all methods.
            \textbf{(a)} shows the evolution of the mean cost during training; after 3 iterations of the alternating loop, our \SDP leads to near-optimal trajectories while all other methods fail (on this task, a cost higher than 200 corresponds to a complete fail).
            As the policy improves, the trajectories collected get more optimal, in a virtuous cycle.
            After 5 iterations, refining policy trajectories takes only 35 TO iterations on average, with the full histogram \textbf{(b)}.
            The action length $T_a$ can be increased to accelerate the policy, inducing a trade-off between policy inference time and TO solving time, as reported in table \textbf{(c)}.
        }
        \label{fig:pend-double}
    \end{center}
\end{figure}

The alternating loop (Alg~\ref{algo:training}, with $\nalgo > 1$) is tested on harder tasks, where the TO solver needs good initial guesses.
After the first training iteration, studied in the previous subsection, new trajectories are collected using the policy to generate initial guesses for the TO solver.
We refer to the process of doing TO on a trajectory produced by a policy as \textit{refining}.
As shown in Fig.~\ref{fig:pend-double}, only \SDP can solve the inverted double pendulum task, while \DP consistently fails to operate torque control, and both \PDDP and \PDDPS strongly diverge.
Not only does \SDP lead to more optimal trajectories compared to TO alone, but it also drastically reduces the solving time.
On this challenging task, without a proper initial guess, the TO solver fails to converge 80\% of the time, while refining from \SDP takes 35 iterations on average.
The optimality of the trajectories produced by \SDP depends on $T_a$, the number of actions applied before replanning.
As shown in Table~\ref{tab:vary_a}, with $T_a=1$, refining a trajectory requires on average 0.22 seconds, compared to 4.1 seconds when solving from scratch (actually, the gap is even wider, as we early stop after 1000 iterations).
$T_a=9$ appears to be a good trade-off, with a policy rollout time of 1.1 seconds and a solving time of 0.56 seconds, but this depends on the GPU and CPU used.  \\[-.5em]

\begin{figure}[t!]
    \begin{center}
        \begin{minipage}{0.87\textwidth}
            \centering
            \includegraphics[width=0.93\columnwidth]{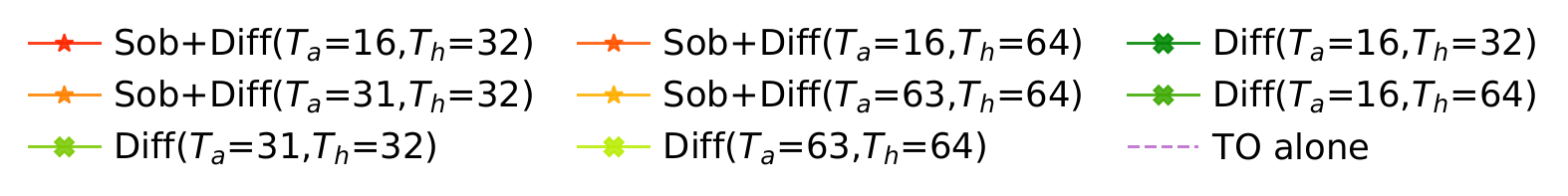}%
            \vspace{-0.2cm}%
            \hspace{0.5em}\phantom{xxx}\underline{\textbf{\UR{1}{0}}} \hspace{6.3em} \underline{\textbf{\UR{1}{1}}}%
            \vspace{0.1cm}%

            \subfigureplot{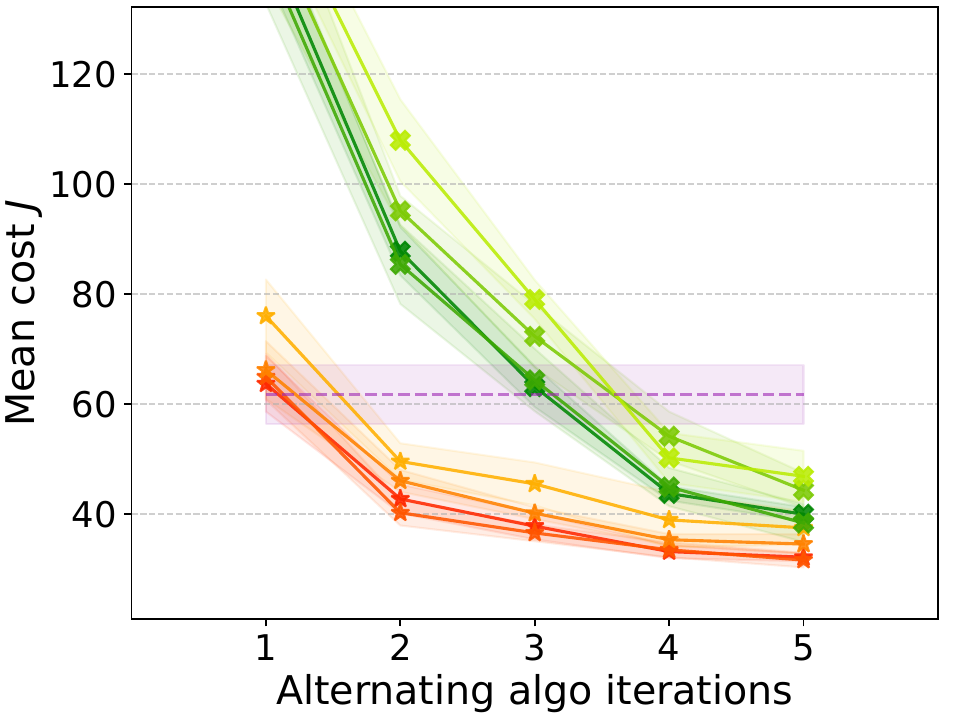}{}{im:h:UR10}{0}
            \hfill
            \subfigureplot{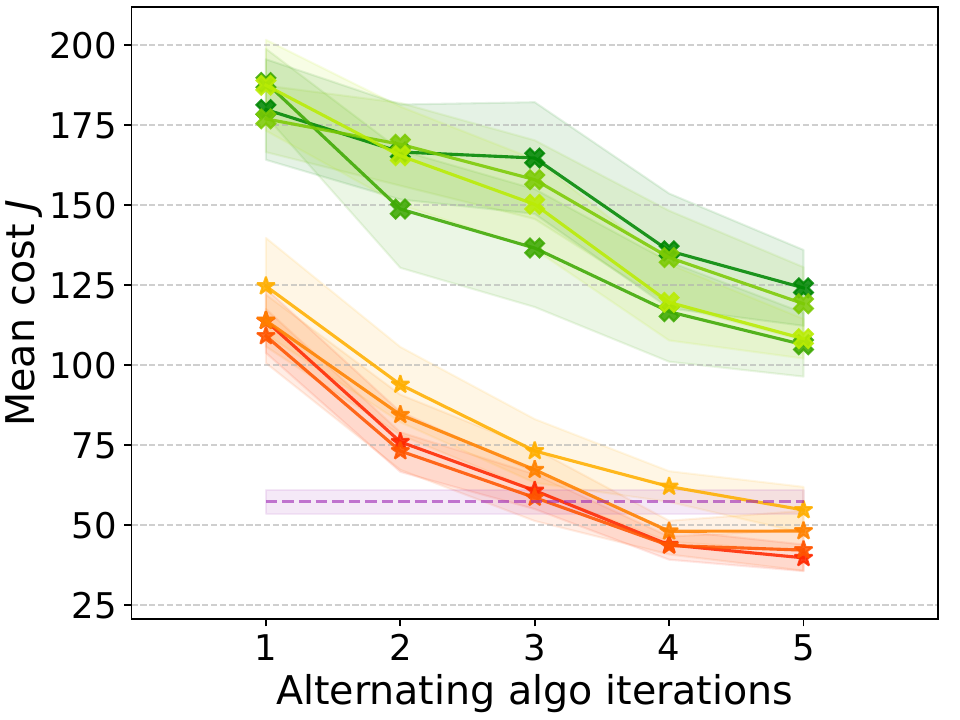}{}{im:h:UR11}{0}

            \vspace{0.10cm}
            \subfigureplot{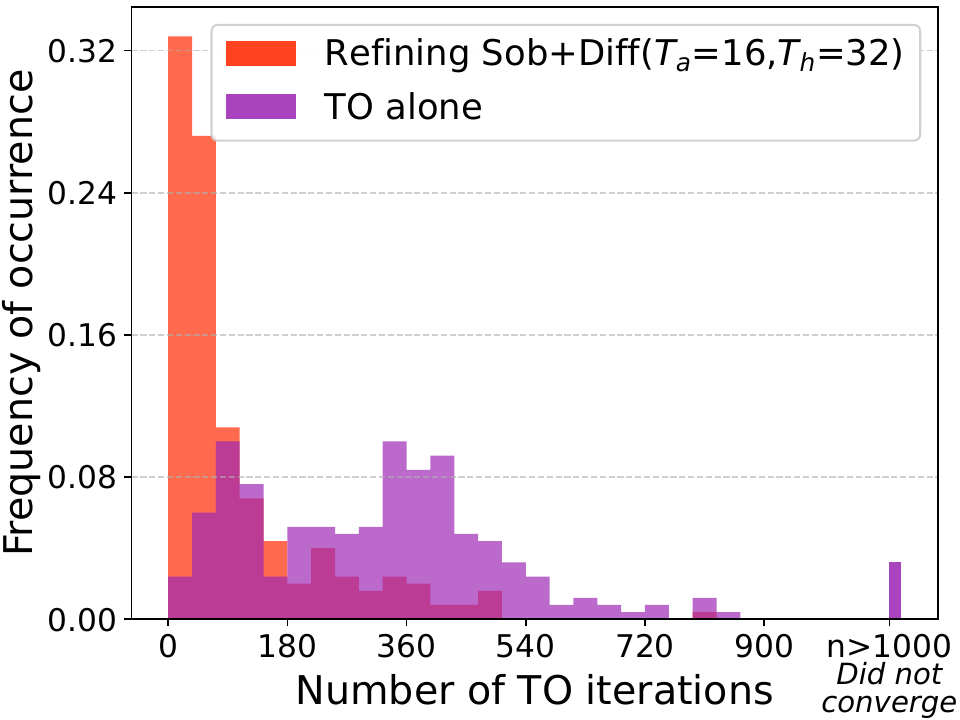}{}{im:hist:UR10}{0}
            \hfill
            \subfigureplot{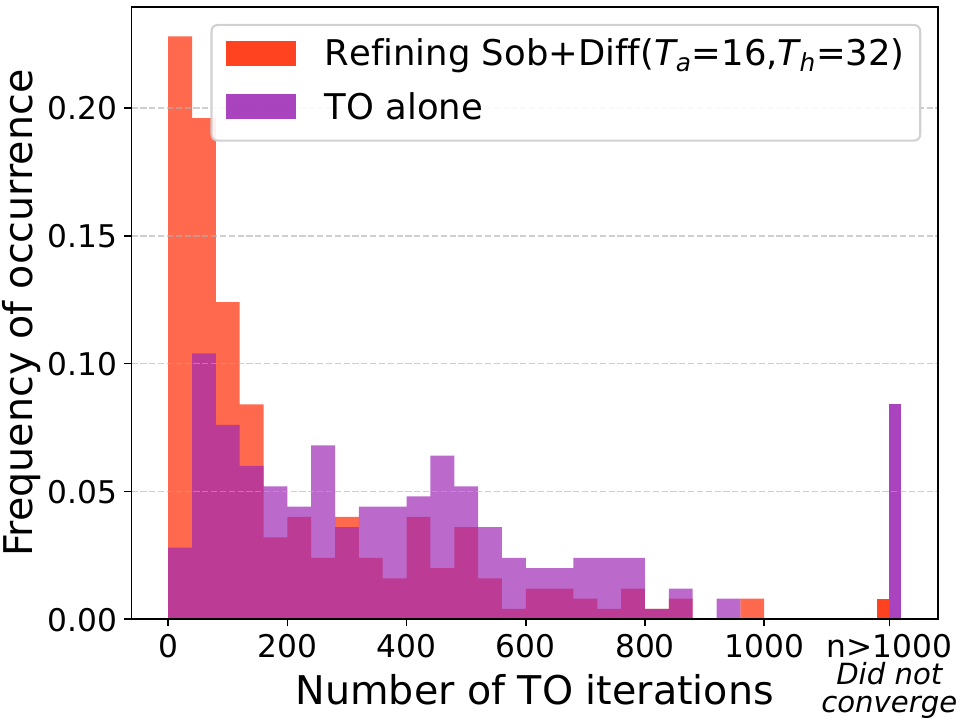}{}{im:hist:UR11}{0}

            \vspace{0.10cm}
            \begin{subfigure}[][][]{0.4\columnwidth}
                \begin{center}%
                    \includegraphics[width=0.95\textwidth,trim={3.0cm 3.5cm 3.0cm 3.5cm},clip]{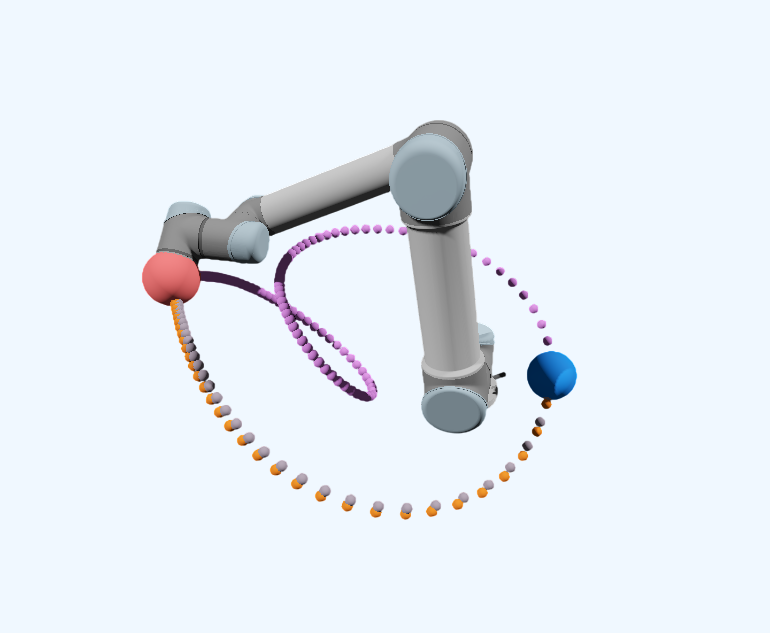}%
                    \vspace{-0.15cm}
                    \caption{}%
                    \label{im:screen:UR10}%
                \end{center}%
            \end{subfigure}%
            \hspace{4.3em}
            \begin{subfigure}[][][]{0.4\columnwidth}
                \hspace{0.09cm}
                \includegraphics[width=0.95\textwidth,trim={1.5cm 2.0cm 1.5cm 2.8cm},clip]{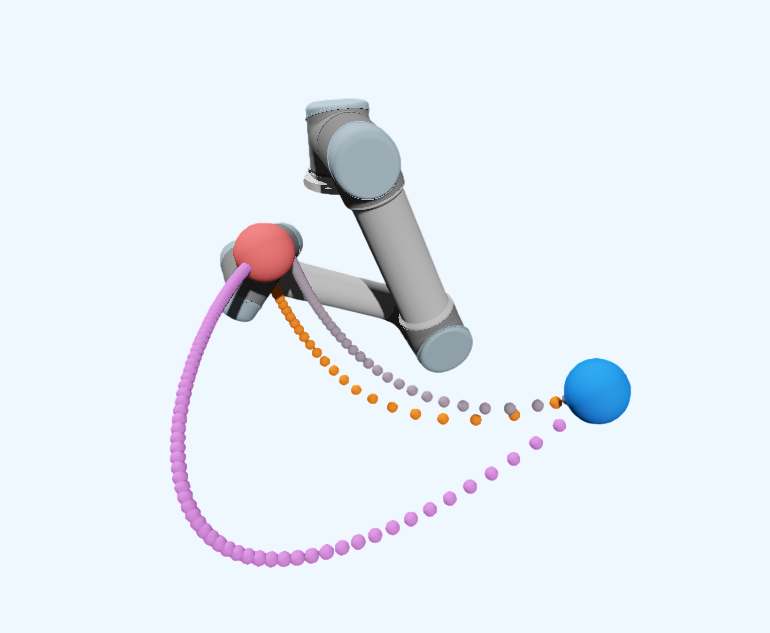}%
                \vspace{-0.15cm}
                \caption{}%
                \label{im:screen:UR11}%
            \end{subfigure}%

            \vspace{-12.1cm}
            \hspace{0.1cm}
            \begin{tikzpicture}
                \draw[dashed] (0,0) -- (0,11.7);
            \end{tikzpicture}
        \end{minipage}

        \caption{\small \textbf{Interplay algorithm on the UR5.}
            \SDPcolor and \DPcolor are evaluated on the challenging \UR{1}{0} (left column), and \UR{1}{1} (right column).
            Methods are tested with varying prediction horizons, $T_h\!=\!32$ and $T_h\!=\!64$, and action lengths $T_a$: 16, 31, and 63 (as $T_a \!\leq T_h - T_o$ and $T_o = 1$).
            For \UR{1}{0}, $\ntraj = 50$ in the first three iterations and 100 after, for \UR{1}{1} $\ntraj$ is doubled, in both cases $\npl=10^4$.
            \SDP solves both tasks, even with a large number of applied actions ($T_a\!=\!63$) for fast inference.
            \textbf{(c)} and \textbf{(d)} report the number of TO iterations taken to converge, with \SDPcolor and \textcolor{oc1}{without}, showing that \SDP helps the solver quickly converge.
            \textbf{(e)} and \textbf{(f)} show an instance of each task, in blue the initial position, in red the goal, in pink trajectories obtained by TO alone (representing trajectories used at the beginning of training), in orange \SDP trajectories, in gray refined ones.
            \SDP trajectories are more direct, avoiding local minima.
        }
        \label{fig:vary-h}
    \end{center}
\end{figure}

\SDP and \DP are further evaluated on the most challenging variants of the UR5 task, using various action lengths.
\SDP consistently shows great control, even with a very long action length $T_a=63$.
While a small $T_a$ may still be preferred for reactivity, we show the resilience of \SDP to large $T_a$, for instance, due to limited computational resources.
On these tasks, warm starting the TO solver with the policy drastically reduces the number of TO iterations and leads to better trajectories, as illustrated by Fig.~\ref{im:screen:UR10} and Fig.~\ref{im:screen:UR11}.

These experiments show that, given a TO solver and tasks it underperforms on, a learned policy can help to find initial guesses.
We do not claim that these tasks are insolvable using dedicated modified iLQR algorithms~\cite{wang2025search}.

\subsection{Tasks involving constraints}
\label{subsec:constrained}

\begin{figure}[h!]
    \begin{center}%
        \rotatebox{90}{\hspace{0.0cm} \large \bf \underline{\OCcolor}}%
        \hspace{0.3cm}%
        \vspace{0.25cm}%
        \includegraphics[angle=270,origin=c,width=0.18\textwidth,trim={2cm 1cm 2cm 1cm},clip]{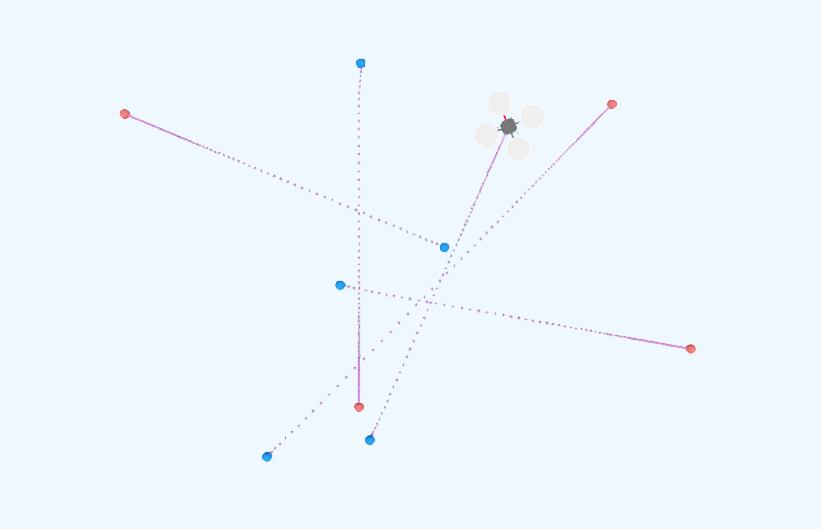}%
        \hspace{0.25cm}%
        \includegraphics[angle=270,origin=c,width=0.18\textwidth,trim={2cm 1cm 2cm 1cm},clip]{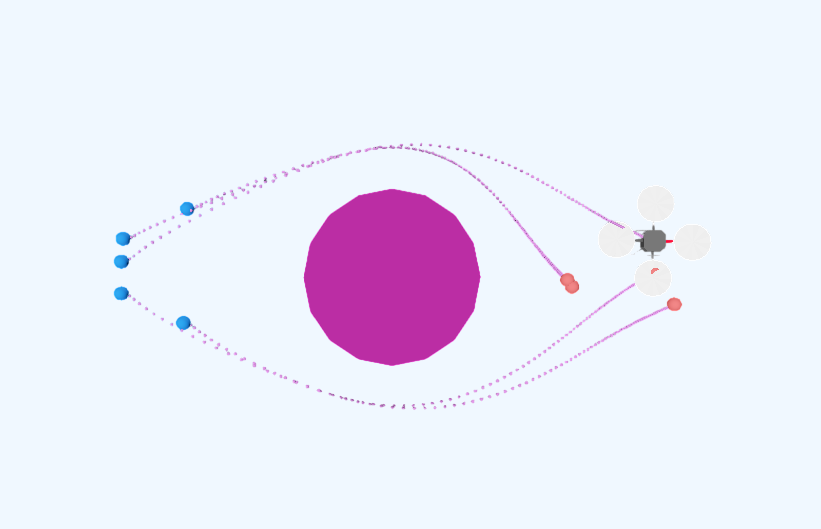}%
        \hspace{0.25cm}%
        \includegraphics[angle=270,origin=c,width=0.18\textwidth,trim={0cm 2.03cm 0cm 2.03cm},clip]{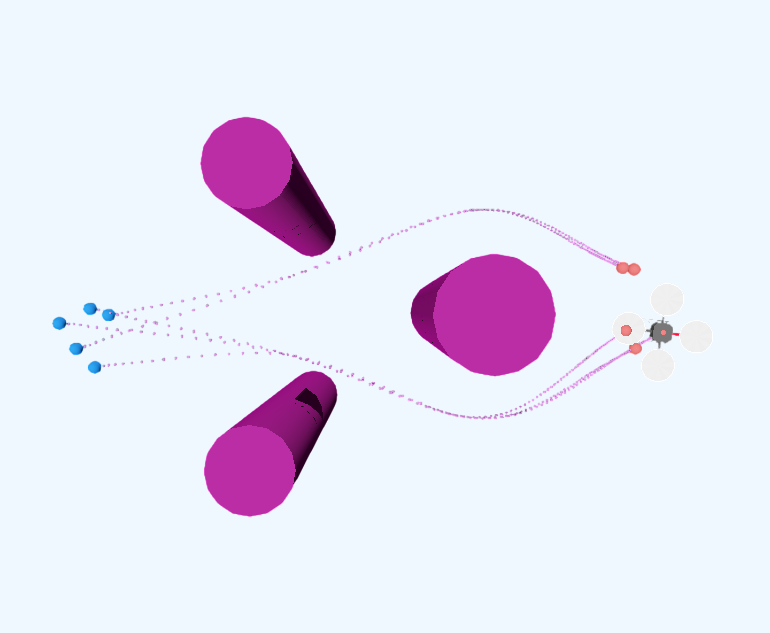}%
        \hspace{0.25cm}%
        \includegraphics[angle=270,origin=c,width=0.18\textwidth,trim={2cm 1cm 2cm 1cm},clip]{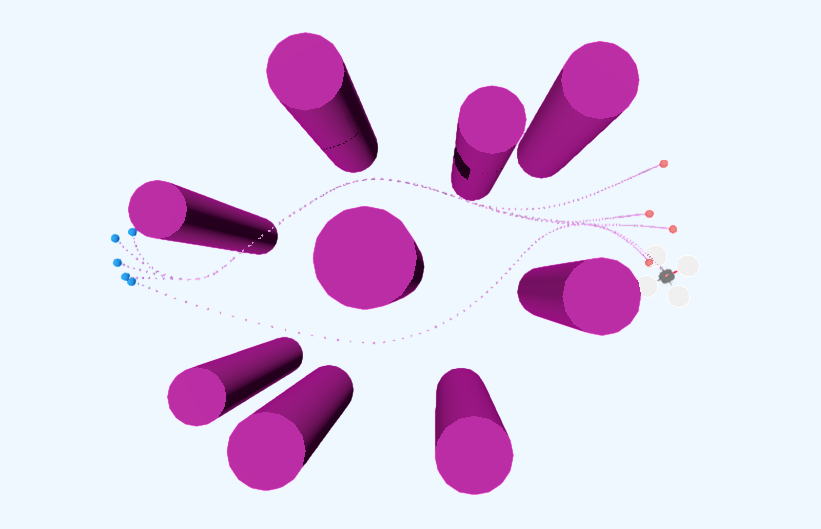}%
        \hspace{0.2cm}%

        \vspace{-0.3em}
        \begin{tikzpicture}
            \draw[dashed] (0,0) -- (11.0,0);
        \end{tikzpicture}
        \vspace{0.3em}

        \rotatebox{90}{\hspace{0.5cm} \large \bf \underline{\DPcolor}}%
        \hspace{0.35cm}%
        \vspace{0.25cm}%
        \includegraphics[angle=270,origin=c,width=0.18\textwidth,trim={2cm 1.5cm 2cm 1.5cm},clip]{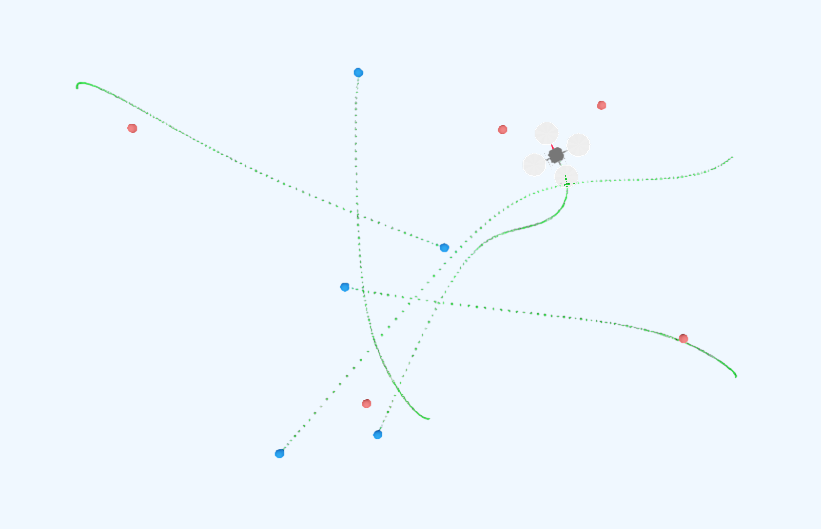}%
        \hspace{0.25cm}%
        \includegraphics[angle=270,origin=c,width=0.18\textwidth,trim={2cm 1.5cm 2cm 1.5cm},clip]{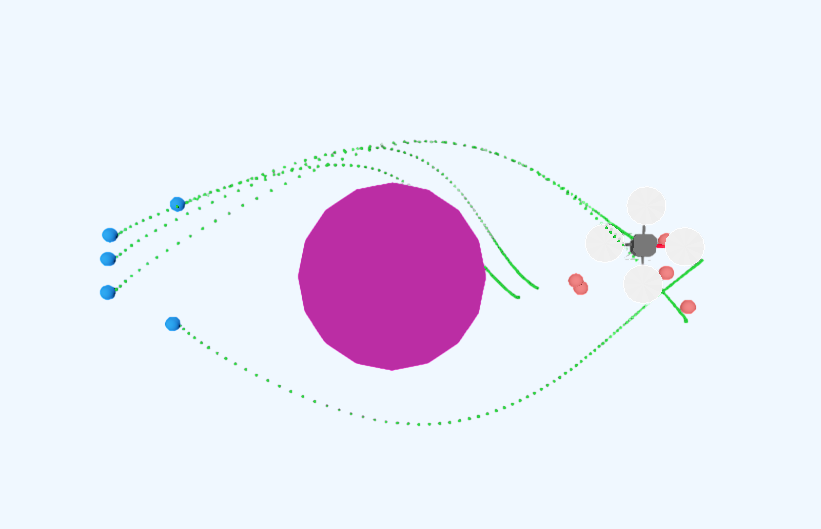}%
        \hspace{0.25cm}%
        \includegraphics[angle=270,origin=c,width=0.18\textwidth,trim={0cm 2.53cm 0cm 2.73cm},clip]{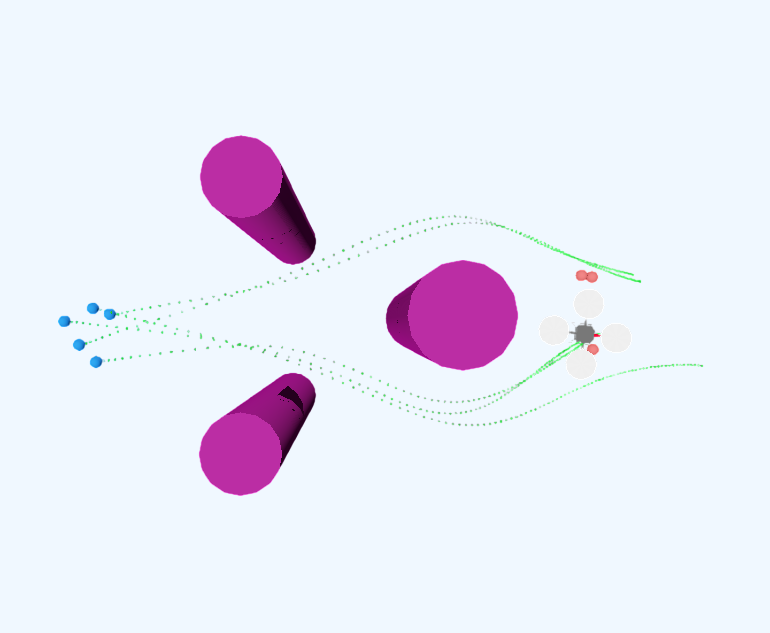}%
        \hspace{0.25cm}%
        \includegraphics[angle=270,origin=c,width=0.18\textwidth,trim={2cm 1.5cm 2cm 1.5cm},clip]{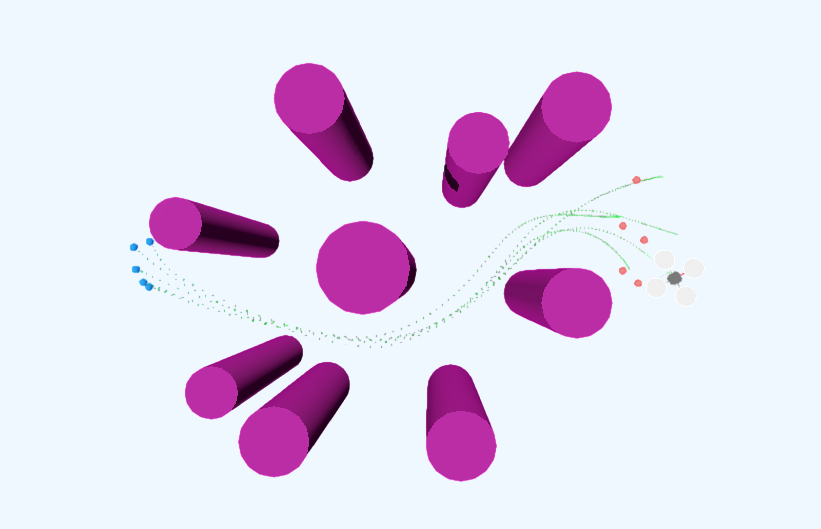}%
        \hspace{0.2cm}%

        \vspace{-0.3em}
        \begin{tikzpicture}
            \draw[dashed] (0,0) -- (11.0,0);
        \end{tikzpicture}
        \vspace{0.3em}

        \rotatebox{90}{\hspace{0.0cm} \large \bf \underline{\SDPcolor}}%
        \hspace{0.35cm}%
        \includegraphics[angle=270,origin=c,width=0.18\textwidth,trim={2cm 1.5cm 2cm 1.5cm},clip]{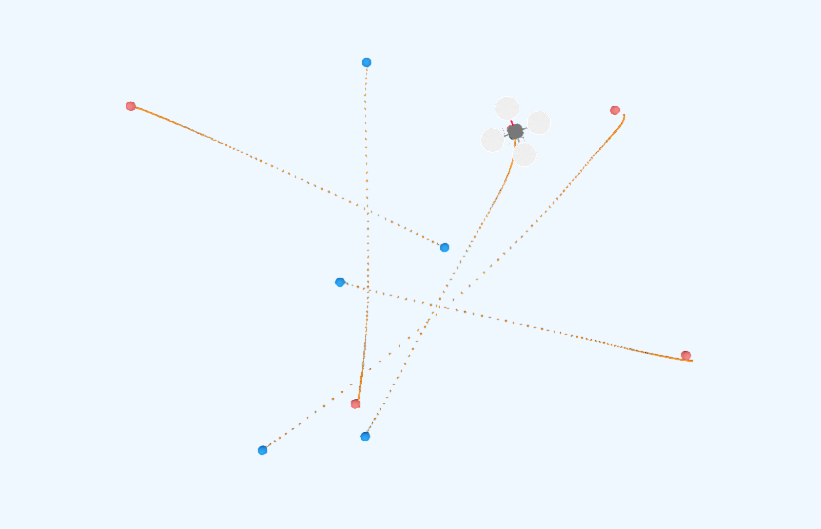}%
        \hspace{0.25cm}%
        \includegraphics[angle=270,origin=c,width=0.18\textwidth,trim={2cm 1.5cm 2cm 1.5cm},clip]{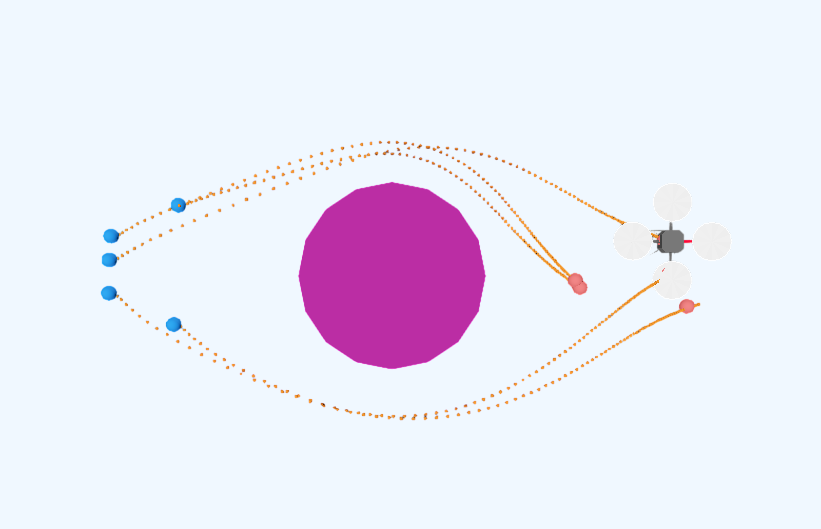}%
        \hspace{0.25cm}%
        \includegraphics[angle=270,origin=c,width=0.18\textwidth,trim={0cm 2.53cm 0cm 2.73cm},clip]{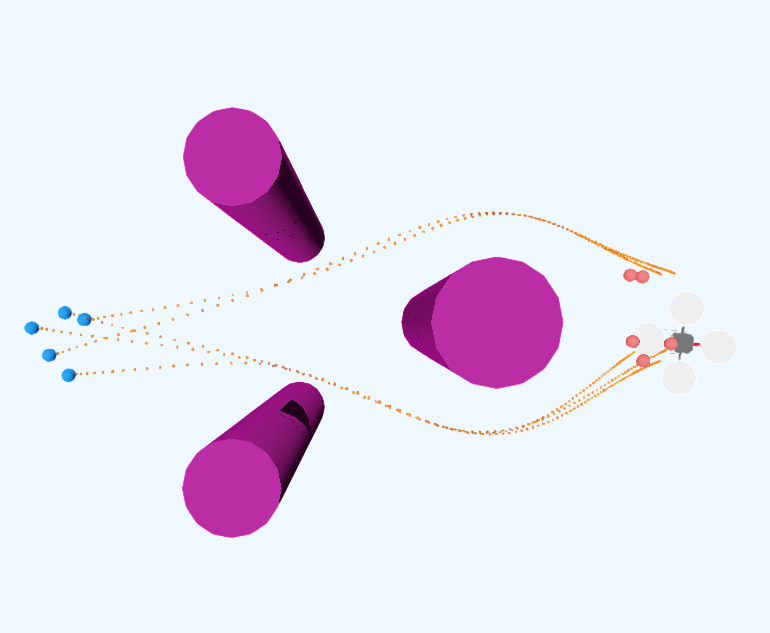}%
        \hspace{0.25cm}%
        \includegraphics[angle=270,origin=c,width=0.18\textwidth,trim={2cm 1.5cm 2cm 1.5cm},clip]{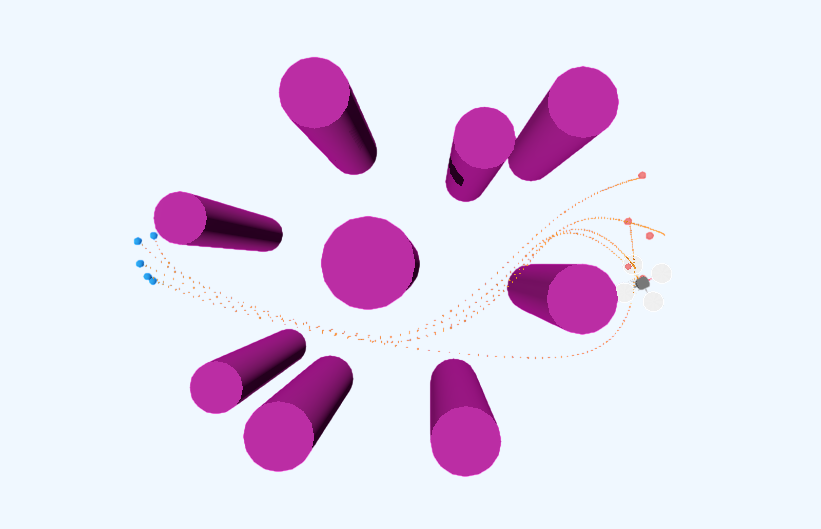}%
        \hspace{0.2cm}%

        \caption{\textbf{Constrained task - Quadrotor with obstacles.}
            \SDPcolor, third row, is compared against \DPcolor, second row, and \textcolor{oc1}{TO} as a reference on the first row.
            We test four variants of the quadrotor task, starting from zero obstacles to nine.
            The obstacles are fixed, for changing number of obstacles one could use a transformer or a recurrent policy.
            Each illustration is done using 5 test instances, hence 5 trajectories (in blue, the initial position, in red, the target one).
            Here we fix $T_h = 32$ and $T_a=16$.
            On the layout with 9 obstacles, we use $\nalgo=5$, $\npl=10^4$, and $\ntraj$ gradually increase at each iteration, from 30 to 100.
            On the other layouts (up to 3 obstacles), $\nalgo=1$ and $\ntraj=50$, no further collecting trajectories.
            For the training time, $\npl=10^4$ when there are no obstacles, then $\npl=2\cdot10^4$ when there are 1 or 3 obstacles.
            The compounding error issue of \DP is visible on the second row: when there are no obstacles, whenever the quadrotor starts to deviate from a straight trajectory, \DP only gets worse.
            When there is one obstacle, if the quadrotor gets too close to the obstacle, \DP is not able to recover, while \SDP adapts smoothly.
            On all trajectories, \SDP exhibits great precision, while \DP nearly never finishes at the target position.
        }
        \label{fig:quadfull}
    \end{center}
\end{figure}

\SDPloss does not have an explicit term to handle constraints; still, we experimented with adding obstacles to the quadrotor task.
On a constrained OCP, reporting an average cost $J$ is harder, because when the constraints are not satisfied, $J$ should be $+\infty$, hiding how much the constraints are violated.
Fig.~\ref{fig:quadfull} illustrates the performances of \SDP and \DP.
In particular, the compounding error issue of \DP is visible: from a small variation, \DP may go straight into an obstacle, thus violating the constraint.
This may be due to $\mathcal{D}$ containing trajectories brushing the obstacles without ever touching them, so touching an obstacle is out of distribution, and \DP gets lost.
On the contrary, we observe that \SDP adapts smoothly.
\SDP exhibits good precision on all trajectories, while \DP rarely finishes at the target position.

On the layout with 9 obstacles, the average solving time of the TO solver without warm-start is 16.3 seconds, while a diffusion policy reduces the solving time to 9.0 seconds, and \SDP further reduces it to 7.8 seconds, \textit{i.e.} by 53\%.

\setlength{\tabcolsep}{4pt}

\section{Discussion and Related Works}
\label{sec:discussion}
We focused on using diffusion policies over trajectories generated by gradient-based TO.
Close to our work, \cite{li2024diffusolve} trains a diffusion model on trajectories obtained from optimal control (DiffuSolve), and, for constrained tasks, a penalty term is added (DiffuSolve+).
As detailed Sec.~\ref{sec:experiments}, in our terms, DiffuSolve is close to \DP with $n_{\textit{algo}}=1$ (no further collecting trajectories).
Adding a similar penalty term to \SDP for constrained problems is interesting future work.

For derivatives matching~\cite{mitchell1992explanation,simard2002transformation,czarnecki2017sobolev,srinivas2018knowledge,pfrommer2022tasil}, TaSIL~\cite{pfrommer2022tasil} is highly relevant as it expands the theory to higher order terms, with bounds in the context of imitation learning and some experiments. Our alternating loop is close to DAgger~\cite{ross2011reduction}, but unlike the assumption of these frameworks, in our case, the TO solver is not a perfect global expert: new trajectories are not just meant to show how to recover from the policy mistakes, the goal is to find better trajectories over time.

Reinforcement learning (RL) is promising to improve the generality of our approach.
On the tasks studied Sec.~\ref{subsec:exp-full}, \SDP worked thanks to the \textit{ArgminCost} term (Alg~\ref{algo:training} Line~\ref{line:argmin-cost}).
This filter was sufficient here, but for more challenging tasks, advanced techniques from offline-RL~\cite{wang2022diffusion,tiofack2025guided}, learning a critic, could be leveraged to better exploit the collected dataset, in particular to filter out low-quality local minima or to estimate the final cost, as in CACTO~\cite{grandesso2023cacto,alboni2024cacto}.
Likewise, curriculum learning could be used to progressively focus on hard task instances.

The goal of this paper is to work along gradient-based TO, by providing initial guesses it reduces the solving time and avoids poor local minima; still, it inherits the limitations of these solvers.
We chose ProxDDP~\cite{jallet2025proxddp} as it directly yields feedback gains, for a first implementation of our method, but we will benefit from developments in this research field.
In particular, leveraging fully differentiable simulators~\cite{lidec2024end} to enable contact-rich tasks.
In particular, \cite{mordatch2014combining} proposed a way to adapt any TO solver to retrieve the feedback gains we need.
While promising, research remains to be conducted as their approach is hard to reproduce.

\section{Conclusion}

We propose a Sobolev approach to train diffusion policies to warm start gradient-based trajectory optimization.
It leverages the first-order information available in this context, for improved sample efficiency, higher policy precision, and longer prediction horizons.
This approach empowers gradient-based TO with an expressive policy to generate initial guesses, reducing the solving time and leading to better solutions.
Future work will focus on contact-rich scenarios, working with new TO solvers, and combining our learning framework with offline RL.
\vspace{-1.em}
\subsubsection{Acknowledgments}
\begin{spacing}{1.1}
    \fontsize{7}{8}\selectfont
    This work has received support from the French government, managed by the National Research Agency, through the INEXACT project (ANR-22-CE33-0007-01) and under the France 2030 program with the references Organic Robotics Program (PEPR O2R) and “PR[AI]RIE-PSAI” (ANR-23-IACL-0008).
    This work was also supported by the European Union through the AGIMUS project (GA no.101070165).
    Views and opinions expressed are those of the author(s) only and do not necessarily reflect those of the European Union or the European Commission. Neither the European Union nor the European Commission can be held responsible for them. \\
    \textbf{Note on LLM usage.} We only used LLMs for basic grammar and spelling corrections.
\end{spacing}

\bibliographystyle{splncs04}
\bibliography{refs}

\end{document}